\pdfoutput=1
\documentclass[]{tokenwave}

\usepackage[utf8]{inputenc}
\usepackage{url}
\usepackage{amsfonts}
\usepackage{amsmath}
\usepackage{nicefrac}
\usepackage{wrapfig}
\usepackage{tikz}
\usepackage{xspace}
\usepackage{array}
\usepackage{pifont}
\usepackage[nointegrals]{wasysym}
\newcommand{\yes}{\leavevmode{\color{twblue}\ding{51}}}
\newcommand{\no}{\leavevmode{\color{black!28}\ding{55}}}
\newcommand{\half}{\leavevmode{\color{twblue}\LEFTcircle}}
\newcommand{\nr}{\leavevmode{\color{black!50}--}}

\newtcolorbox{taskfile}[1]{enhanced, breakable, colback=white, colframe=black!18, boxrule=0.4pt,
  arc=2pt, left=9pt, right=9pt, top=5pt, bottom=6pt, colbacktitle=twbg, coltitle=black!70,
  fonttitle=\ttfamily\footnotesize, title=#1, fontupper=\small\raggedright, before skip=6pt, after skip=10pt}
\newcommand{\dochead}[2]{\par\smallskip\noindent{\sffamily\bfseries\footnotesize #1}\hfill
  {\sffamily\scriptsize\color{twblue}#2}\par\nopagebreak}
\newcommand{\docsub}[1]{\par\noindent{\sffamily\footnotesize #1}\par\nopagebreak}
\newcommand{\docskip}{\par\noindent{\color{black!45}[\,\ldots\,]}\par}
\newenvironment{docitems}{\begin{itemize}\setlength{\itemsep}{0pt}\setlength{\parskip}{0pt}%
  \setlength{\topsep}{2pt}}{\end{itemize}}
\usepackage{longtable}
\usetikzlibrary{positioning, arrows.meta, fit, backgrounds, calc}

\crefname{figure}{Figure}{Figures}
\Crefname{figure}{Figure}{Figures}
\crefname{table}{Table}{Tables}
\Crefname{table}{Table}{Tables}
\crefname{section}{Section}{Sections}
\Crefname{section}{Section}{Sections}
\crefname{subsection}{Section}{Sections}
\Crefname{subsection}{Section}{Sections}
\crefname{equation}{Equation}{Equations}
\Crefname{equation}{Equation}{Equations}
\crefname{appendix}{Appendix}{Appendices}
\Crefname{appendix}{Appendix}{Appendices}
\crefname{subappendix}{Appendix}{Appendices}
\Crefname{subappendix}{Appendix}{Appendices}

\titleformat*{\paragraph}{\sffamily\bfseries}

\newcommand{\finding}[1]{{\rmfamily\bfseries #1}}

\newcommand{\syslogo}[1]{\raisebox{-0.18em}{\includegraphics[height=1em]{figures/logos/#1.pdf}}}
\newlength{\langlogowidth}
\newcommand{\langicon}[1]{\includegraphics[height=1em,width=1.75em,keepaspectratio]{figures/logos/lang/#1.pdf}}
\newcommand{\langlogo}[1]{\makebox[\langlogowidth][c]{\raisebox{\dimexpr0.36em-0.5\height\relax}{\langicon{#1}}}}
\newcommand{\langspace}{\makebox[\langlogowidth]{}}
\newcommand{\pkdef}[2]{\expandafter\def\csname pk@#1\endcsname{#2}}
\newcommand{\pk}[1]{\ifcsname pk@#1\endcsname\csname pk@#1\endcsname\else\textcolor{red}{[#1?]}\fi}
\pkdef{audit-high}{92.2}
\pkdef{audit-low}{90.7}
\pkdef{audit-tasks}{129}
\pkdef{audit-tests}{172,448}
\pkdef{b-grok-completion-overclaim}{49}
\pkdef{b-max-full-spec-read-first}{26}
\pkdef{b-max-interface-audit}{99}
\pkdef{b-max-spec-example-checks}{100}
\pkdef{b-min-full-spec-read-first}{1}
\pkdef{b-min-interface-audit}{71}
\pkdef{b-min-spec-example-checks}{86}
\pkdef{b-opus-recalled-upstream-tests}{32}
\pkdef{base-max}{18.1}
\pkdef{base-min}{14.5}
\pkdef{collapsed-runs-pct}{4.7}
\pkdef{collapsed-share}{45}
\pkdef{commits-2026}{145}
\pkdef{comp-max}{7}
\pkdef{core-graded}{1,335}
\pkdef{core-share}{13.0}
\pkdef{created-2026}{44}
\pkdef{deepseek-base-nospec}{12.4}
\pkdef{deepseek-core-gap}{15.9}
\pkdef{deepseek-core-gap-ctrl}{6.5}
\pkdef{deepseek-corefar}{25.5}
\pkdef{deepseek-depth-single}{21.7}
\pkdef{deepseek-depth-slope}{22.4}
\pkdef{deepseek-doubling}{10.0}
\pkdef{deepseek-far}{37.7}
\pkdef{deepseek-hit10h}{6}
\pkdef{deepseek-not-run}{2.8}
\pkdef{deepseek-old-names}{4}
\pkdef{deepseek-said-overlap}{0.07}
\pkdef{features-all}{11,124}
\pkdef{features-graded}{10,235}
\pkdef{fov-did}{+0.7}
\pkdef{fov-did-hi}{+3.8}
\pkdef{fov-did-lo}{$-$2.5}
\pkdef{fov-features}{10,228}
\pkdef{fov-high-old}{76}
\pkdef{fov-opus-high}{60.7}
\pkdef{fov-opus-low}{63.0}
\pkdef{fov-sol-high}{45.5}
\pkdef{fov-sol-low}{48.5}
\pkdef{fov-tasks}{148}
\pkdef{gap-max}{35.8}
\pkdef{gap-min}{25.5}
\pkdef{grid-name}{pdm}
\pkdef{grok-base-nospec}{14.6}
\pkdef{grok-core-gap}{10.5}
\pkdef{grok-core-gap-ctrl}{4.7}
\pkdef{grok-corefar}{14.0}
\pkdef{grok-depth-single}{30.2}
\pkdef{grok-depth-slope}{23.6}
\pkdef{grok-doubling}{6.2}
\pkdef{grok-far}{22.5}
\pkdef{grok-last-smoke}{84}
\pkdef{grok-no-test-runs}{53}
\pkdef{grok-not-run}{27.9}
\pkdef{grok-old-names}{1}
\pkdef{grok-said-overlap}{0.02}
\pkdef{hi95-far-gap}{16.2}
\pkdef{hi95-n}{32}
\pkdef{hi95-t-gap}{1.7}
\pkdef{hit10h-total}{13}
\pkdef{lead-change}{$-$14.9}
\pkdef{lead-change-hi}{$-$5.8}
\pkdef{lead-change-lo}{$-$23.5}
\pkdef{lead-java}{17.8}
\pkdef{lead-new}{7.3}
\pkdef{lead-old}{22.2}
\pkdef{lead-python}{19.3}
\pkdef{lead-rest-max}{9.3}
\pkdef{lead-rest-min}{6.9}
\pkdef{lead-y2021}{19.5}
\pkdef{lead-y2024}{5.4}
\pkdef{lic-none}{6}
\pkdef{lic-permissive}{119}
\pkdef{lic-restrictive}{3}
\pkdef{lic-strong-copyleft}{15}
\pkdef{lic-weak-copyleft}{7}
\pkdef{max-feature-tests}{11,851}
\pkdef{max-groups}{35}
\pkdef{max-tokens-m}{1.26}
\pkdef{med-ac}{490}
\pkdef{med-features}{44}
\pkdef{med-fr}{1,042}
\pkdef{med-graded-tests}{750}
\pkdef{med-groups}{9}
\pkdef{med-spec-features}{51}
\pkdef{med-src-lines-k}{74}
\pkdef{med-tests-per-feature}{11}
\pkdef{med-tokens-k}{180}
\pkdef{med-top-feature-share}{39.3}
\pkdef{min-groups}{2}
\pkdef{modelaudit-overlap}{1.2}
\pkdef{modelaudit-tests}{13}
\pkdef{new-n}{44}
\pkdef{none-share}{29.0}
\pkdef{old-n}{59}
\pkdef{opus-base-nospec}{10.9}
\pkdef{opus-common}{150}
\pkdef{opus-core-gap}{11.6}
\pkdef{opus-core-gap-ctrl}{3.0}
\pkdef{opus-corefar}{51.8}
\pkdef{opus-delta-abs}{14.6}
\pkdef{opus-delta-hi}{+18.4}
\pkdef{opus-delta-lo}{+10.9}
\pkdef{opus-depth-single}{12.0}
\pkdef{opus-depth-slope}{13.2}
\pkdef{opus-doubling}{9.1}
\pkdef{opus-far}{61.7}
\pkdef{opus-grams-new}{28.5}
\pkdef{opus-grams-old}{58.9}
\pkdef{opus-hit10h}{7}
\pkdef{opus-old-names}{20}
\pkdef{opus-old-overlap}{50.4}
\pkdef{opus-resolve}{7}
\pkdef{opus-said-overlap}{0.63}
\pkdef{opus-said-runs}{144}
\pkdef{opus-t}{87.2}
\pkdef{opus-w-out}{2.8 million}
\pkdef{opus-w-subagents}{25}
\pkdef{opus-wins}{105}
\pkdef{ord-edge-max}{70}
\pkdef{ord-edge-min}{61}
\pkdef{ord-far-max}{0.06}
\pkdef{ord-far-min}{$-$0.06}
\pkdef{ord-first-max}{4}
\pkdef{ord-first-min}{2}
\pkdef{ord-gated-max}{3}
\pkdef{ord-last-max}{11}
\pkdef{ord-last-min}{6}
\pkdef{ord-null-max}{49}
\pkdef{ord-null-min}{47}
\pkdef{ord-read-max}{0.67}
\pkdef{ord-read-min}{0.36}
\pkdef{ord-started-max}{85}
\pkdef{ord-started-min}{81}
\pkdef{ord-unread-max}{41}
\pkdef{ord-unread-min}{9}
\pkdef{orig-cpl}{39}
\pkdef{orig-unformatted}{1.6}
\pkdef{overclaim-rest-max}{29}
\pkdef{overclaim-rest-min}{1}
\pkdef{placed-entries}{12,770}
\pkdef{placed-nontest-src}{708}
\pkdef{placed-nontest-src-tasks}{41}
\pkdef{placed-tasks}{109}
\pkdef{q1-src-lines-k}{34}
\pkdef{q3-src-lines-k}{126}
\pkdef{reach-mean}{2.32}
\pkdef{reach-one}{48.7}
\pkdef{resolve-total}{7}
\pkdef{runs-lead-total}{600}
\pkdef{sample-ac}{15}
\pkdef{sample-fr}{20}
\pkdef{sol-base}{14.5}
\pkdef{sol-base-nospec}{1.9}
\pkdef{sol-base-raw}{2.2}
\pkdef{sol-core-gap}{12.4}
\pkdef{sol-core-gap-ctrl}{2.4}
\pkdef{sol-corefar}{37.4}
\pkdef{sol-cpl}{63}
\pkdef{sol-depth-single}{15.9}
\pkdef{sol-depth-slope}{17.2}
\pkdef{sol-doubling}{10.5}
\pkdef{sol-far}{47.1}
\pkdef{sol-not-run}{4.5}
\pkdef{sol-w-out}{308 thousand}
\pkdef{sol-w-subagents}{3}
\pkdef{sol5-always}{27.9}
\pkdef{sol5-best1}{47.7}
\pkdef{sol5-best5}{57.5}
\pkdef{sol5-collapsed}{35}
\pkdef{sol5-never}{35.2}
\pkdef{sol5-out-hi}{17.6}
\pkdef{sol5-out-hi-nc}{4.5}
\pkdef{sol5-out-lo}{4.3}
\pkdef{sol5-out-lo-nc}{$-$1.8}
\pkdef{sol5-out-slope}{11.4}
\pkdef{sol5-out-slope-nc}{1.1}
\pkdef{sol5-run-max}{49.1}
\pkdef{sol5-run-min}{47.0}
\pkdef{sol5-runs}{750}
\pkdef{sol5-sometimes}{36.9}
\pkdef{spec-lines-max}{31.1}
\pkdef{spec-lines-min}{17.5}
\pkdef{start-java}{1}
\pkdef{start-lines}{17,069}
\pkdef{start-rust}{15}
\pkdef{start-share}{0.88}
\pkdef{start-tasks}{16}
\pkdef{tasks}{150}
\pkdef{tasks-final}{150}
\pkdef{types}{14}
\pkdef{types-same-order}{13}

\title{NL2Repo-Pro: Benchmarking Multilingual Production-Scale Library Development Feature by Feature}

\author[1,2]{Heli Qi}
\author[3]{Minghao Liu}
\author[4]{Jiarui Liu}
\author[5]{Jingzhe Ding}
\author[6]{Qingcheng Zeng}
\author[7]{Weijia Zhang}
\author[4]{Yunze Xiao}
\author[8]{Yuchen Wu}
\author[5]{Bing Zhao}
\author[9]{Jen-Tse Huang}
\author[1]{Stephen Huang}
\author[10]{Zhijing Jin}
\author[1]{Eli Cheung}
\author[4]{Mona Diab}
\author[1,3]{Weihao Xuan}

\affiliation[1]{TokenWave AI\protect\textsuperscript{\protect\textdagger}}
\affiliation[2]{Waseda University}
\affiliation[3]{The University of Tokyo}
\affiliation[4]{Carnegie Mellon University}
\affiliation[5]{Independent Researcher}
\affiliation[6]{Northwestern University}
\affiliation[7]{Yale University}
\affiliation[8]{New York University}
\affiliation[9]{Johns Hopkins University}
\affiliation[10]{University of Toronto}
\newcommand{\affillogoheight}{0.8cm}
\newcommand{\affillogo}[2]{%
  \IfFileExists{figures/logos/affil/#1.pdf}%
    {\includegraphics[height=\affillogoheight]{figures/logos/affil/#1.pdf}}%
    {\IfFileExists{figures/logos/affil/#1.png}%
      {\includegraphics[height=\affillogoheight]{figures/logos/affil/#1.png}}%
      {\tikz[baseline=(b.south)]\node[name=b, draw=twnavy!35, fill=white, rounded corners=2pt,
         minimum width=\affillogoheight, minimum height=\affillogoheight, inner sep=1pt,
         font=\sffamily\bfseries\fontsize{5.5}{6}\selectfont, text=twnavy!70] {#2};}}}
\renewcommand{\twpartnerlogos}{%
  \affillogo{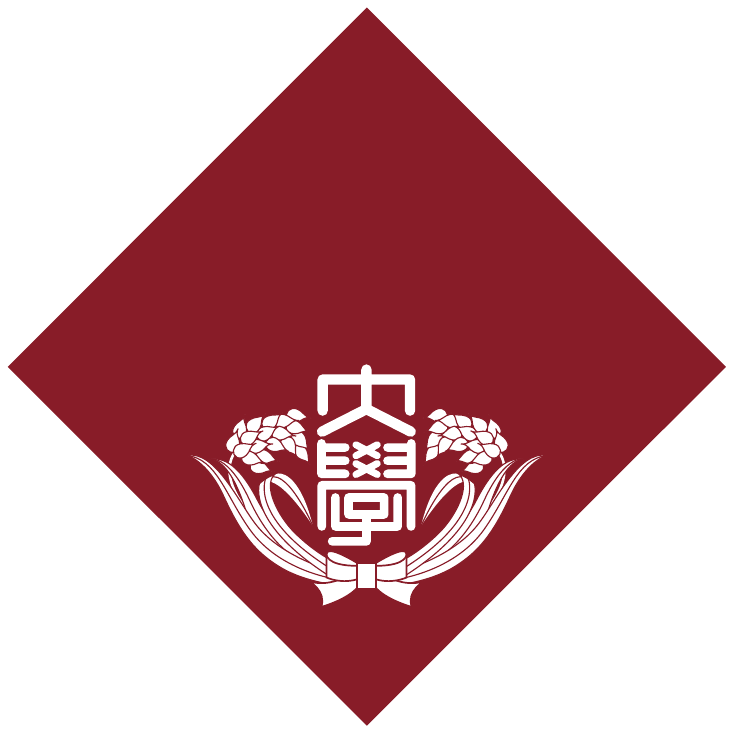}{Waseda}\hspace{0.22cm}%
  \affillogo{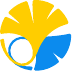}{UTokyo}\hspace{0.22cm}%
  \affillogo{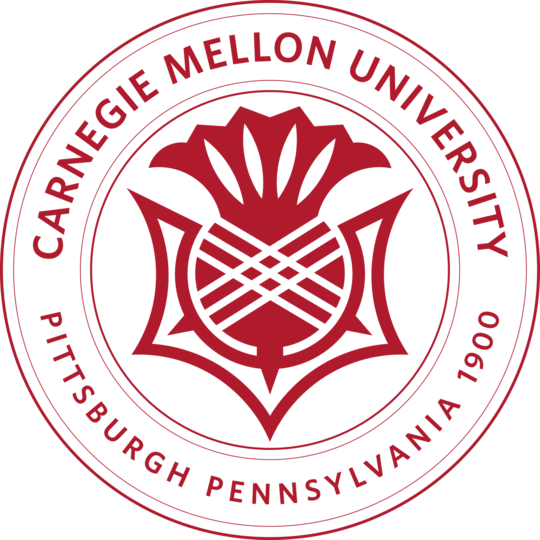}{CMU}\hspace{0.22cm}%
  \affillogo{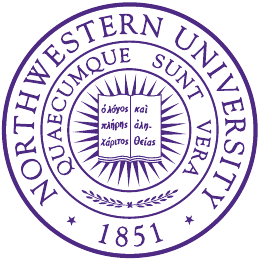}{NWU}\hspace{0.22cm}%
  \affillogo{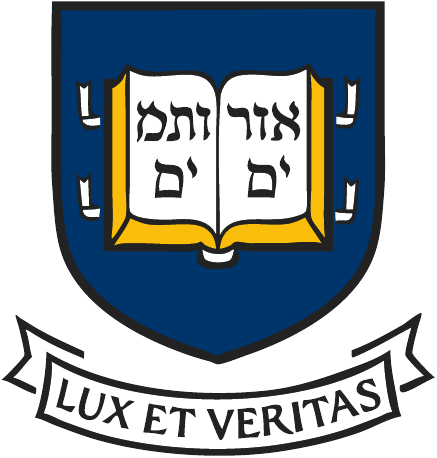}{Yale}\hspace{0.22cm}%
  \affillogo{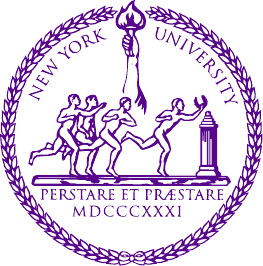}{NYU}\hspace{0.22cm}%
  \affillogo{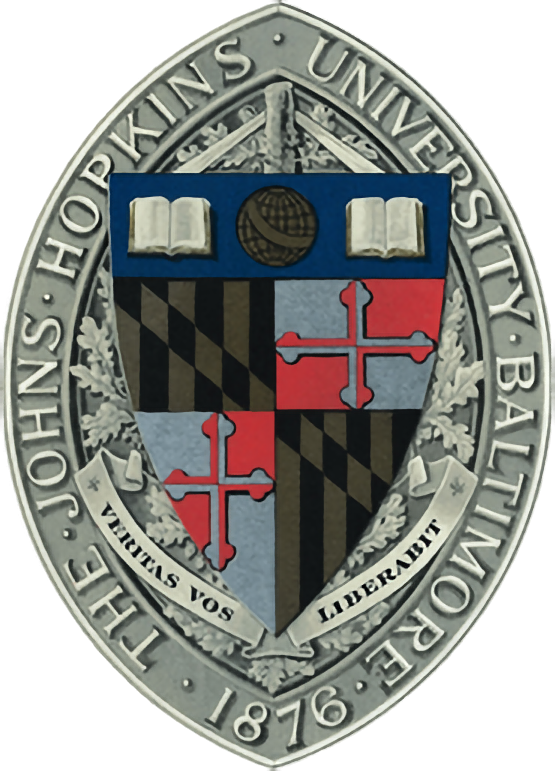}{JHU}\hspace{0.22cm}%
  \affillogo{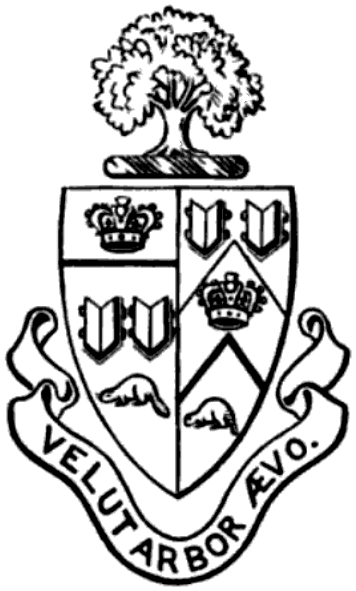}{UofT}}

\abstract{%
Coding agents have progressed from writing single functions to resolving issues in real repositories and, most recently, to building entire repositories from natural-language requirements. As these repositories grow to the size of a production library, benchmarks face two challenges: specifying requirements that depend on one another, and measuring how much of a delivery works, which a whole-task verdict misses and a test pass rate can overstate. We present NL2Repo-Pro, a benchmark of \pk{tasks-final} tasks built from real libraries with a median of \pk{med-src-lines-k} thousand lines of code, spanning five programming languages and \pk{types} project types. Each task specifies a library through an overview document, which arranges its features in dependency layers, and a specification for each feature. From these documents, the agent builds the library from scratch, without network access and without seeing the original repository's tests. These tests then grade the delivery feature by feature: a feature is accepted when every test that directly calls its code passes, and the main score, the Feature Acceptance Rate (FAR), is the share of features accepted. Of the four frontier systems we evaluate, Claude Opus 5.5 is the strongest. It passes \pk{opus-t}\% of the tests but has a FAR of \pk{opus-far}\%, and only \pk{opus-resolve} of its \pk{tasks} libraries have every feature accepted. The other three systems reach a FAR of \pk{grok-far}\% to \pk{sol-far}\%. The ranking does not rest on recall: it holds on the \pk{new-n} repositories created in 2026, where no system writes even 1\% of the hidden test names. FAR also separates systems with nearly equal test pass rates: on the \pk{hi95-n} tasks where Claude Opus 5.5 and GPT-6.1 Sol both pass at least 95\% of the tests, their test pass rates differ by \pk{hi95-t-gap} points and their FAR by \pk{hi95-far-gap}. Even the strongest system builds most of a production library but rarely a complete one, and getting every feature fully right remains an open challenge for coding agents.
}

\twinfo[Website]{\url{https://tokenwave-ai.github.io/NL2Repo-Pro/}}
\twinfo[Code]{\url{https://github.com/TokenWave-AI/NL2Repo-Pro}}
\twinfo[Data]{\url{https://huggingface.co/datasets/Tokenwave-AI/NL2Repo-Pro}}

\begin{document}

\maketitle
\begingroup\renewcommand{\thefootnote}{\textdagger}\footnotetext{Located in Palo Alto, United States.}\endgroup

\section{Introduction}
\label{sec:intro}

Coding agents and their benchmarks have co-evolved: as agents grew more capable, benchmarks raised the unit of work from single functions \citep{chen2021evaluatinglargelanguagemodels,austin2021programsynthesislargelanguage} to issue fixes in real repositories \citep{jimenez2024swebenchlanguagemodelsresolve} and then to whole repositories written from a specification \citep{ding2026e2eswe,ding2026nl2repo}. Agents keep improving at a steady pace: the length of tasks that frontier agents complete with 50\% success, measured in expert time, has doubled roughly every seven months since 2019 and reached about twelve hours for publicly available models in early 2026 \citep{kwa2025measuringaiabilitycomplete,metr2026riskreport}. Yet most benchmarks that build a repository from scratch and report their size still target programs of under ten thousand lines (\Cref{tab:related}). The next unit of work is a complete library of the size used in production, and building one raises problems that smaller tasks do not.

Evaluating a unit of work this large is a challenge in itself. The question shifts from whether a delivery works to how much of it works, and neither of the usual scores answers it well. The resolve rate of issue-level benchmarks rests on a whole-task verdict, which credits a run only when every test passes, and at this scale almost no run does: on the benchmark we introduce below, only \pk{resolve-total} of the \pk{runs-lead-total} runs of four frontier systems pass every test. The test pass rate gives partial credit, but it lets the features with the most tests decide the outcome. To our knowledge, no benchmark that builds production-scale repositories from scratch scores a delivery by the features that the original tests accept. Additionally, the requirements of such a library no longer fit in one document. Requirements practice keeps a product overview apart from detailed requirements, which can be organized by feature \citep{wiegers2013software}, and the features of a production library build on one another, so an agent must keep the whole library in view across many documents and decide in what order to build. A run also becomes long enough that a score alone does not describe it: a single run can process hundreds of millions of tokens, and systems differ widely in how many resources they use and in how they organize the work.

\begin{figure}[t]
  \centering
  \includegraphics[width=\linewidth]{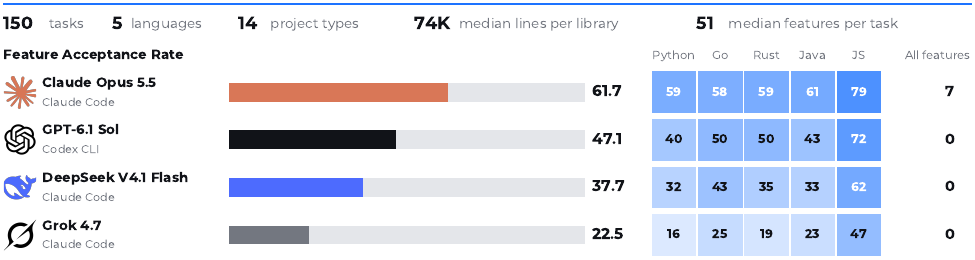}
  \caption{NL2Repo-Pro at a glance. Left: the Feature Acceptance Rate (FAR) of each system, in percent. Right: FAR by language, and the number of tasks with every feature accepted.}
  \label{fig:teaser}
\end{figure}

\begin{figure}[t]
  \centering
  \includegraphics[width=\linewidth]{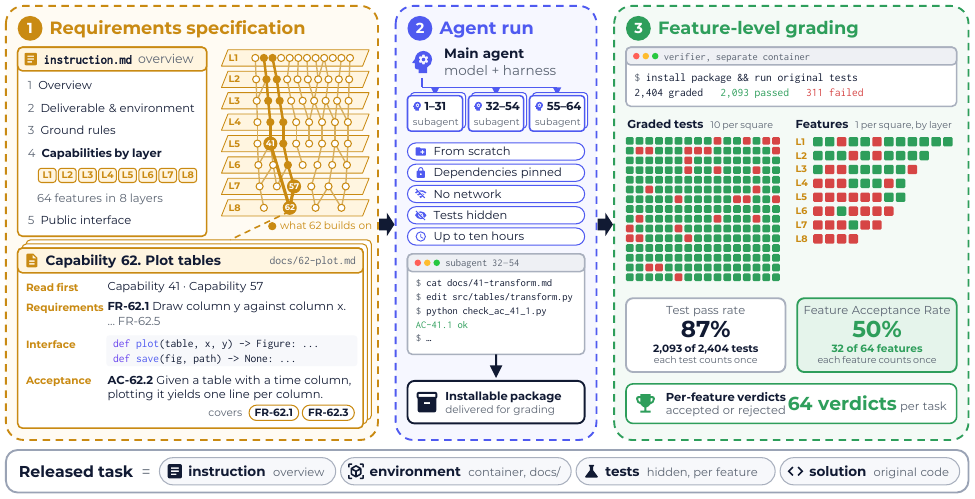}
  \caption{A task and its grading. \textbf{(1)} The agent reads a requirements specification:
  an overview document that lists the features by dependency layer, and one specification per feature
  with requirements, an interface and acceptance criteria. Task files label each feature as \texttt{Capability $N$}.
  In the graph each dot is a feature, L1 to L8 are the dependency layers, and a line joins a feature to one it uses.
  Feature 62, whose specification is shown, and the features it builds on are highlighted.
  \textbf{(2)} A main agent and its subagents build the library from scratch, offline and without the tests,
  and deliver an installable package. \textbf{(3)} The original tests grade the package in a separate container.
  A test counts for every feature it calls directly, and a feature is accepted only when all of its tests pass.
  Here 87\% of 2,404 graded tests pass, yet only 32 of 64 features are accepted (FAR = 50\%).
  The bottom row lists what a released task contains. All content is illustrative.}
  \label{fig:task-form}
\end{figure}

In our earlier work, NL2Repo-Bench \citep{ding2026nl2repo}, we asked agents to build Python libraries of a few thousand lines, each from a single requirements document, and scored each delivery by its test pass rate. In this paper, we take this task to production scale and introduce NL2Repo-Pro, a benchmark of \pk{tasks-final} tasks built from real repositories in Python, Go, Rust, Java and JavaScript, with a median of \pk{med-src-lines-k} thousand lines of code in the original library. At this scale, both the specification and the scoring are organized by feature. A feature is a unit of functionality that can be specified and accepted on its own. The specification becomes a set of documents: an overview document describes the library and groups its features into dependency layers, and each feature has its own specification with requirements, an interface and acceptance criteria, for a median of \pk{med-spec-features} features in \pk{med-groups} layers. All documents are available from the start and no stage gate tells the agent when to move on, so the order of work is the agent's own decision. The agent builds the library from scratch, without network access or visible tests. The original repository's tests then grade the delivery, and every graded test is attributed to the features whose code it calls directly. The main score, the Feature Acceptance Rate (FAR), is the share of features whose graded tests all pass, so it shows how much of a library works. A task thus yields one verdict per feature instead of one for the whole task, and these verdicts can be read layer by layer (\Cref{fig:task-form}). We evaluate four agentic systems, each a model paired with an agent harness (\Cref{fig:teaser}), and report three findings.

\finding{First, frontier systems rarely deliver a fully working library, even when they pass most of its tests.} Claude Opus 5.5, the top system, passes \pk{opus-t}\% of the graded tests, yet only \pk{opus-far}\% of its features are accepted, and it has every feature accepted on just \pk{opus-resolve} of the \pk{tasks} tasks. For every system, the test pass rate overstates how much of a library works. Per-feature verdicts show more than either score: read by dependency layer, they show that every system has the most features accepted in the first layers (\Cref{sec:main-results}).

\finding{Second, in terms of behavior, systems use the layered specification as a guide to dependencies rather than as a fixed schedule.} They usually do not finish one layer before starting the next, and work on several layers at once. Even so, their order follows the dependencies: a feature tends to be built before the features that depend on it, more often than a random order would give. With the tests hidden, every system also turns to the specification to check its work, comparing its code against the acceptance criteria in most of the runs where this can be judged (\Cref{sec:work}).

\finding{Third, results show a temporal effect: newer repositories bring fewer signs of recall and a smaller lead for the top system.} On about a third of the repositories from 2020 or earlier, Claude Opus 5.5 writes at least 1\% of the names of the hidden tests, which a task never shows, while on those created in 2026 no system does. Its lead over GPT-6.1 Sol also shrinks from \pk{lead-old} points on the oldest repositories to \pk{lead-new} on the newest (\Cref{sec:recall}). To sum up, our contributions are as follows.
\begin{enumerate}
  \item We build NL2Repo-Pro, a benchmark of \pk{tasks-final} production-scale library-construction tasks in five languages, and ensure its quality by having software engineers review every task at each stage of construction (\Cref{sec:construction,sec:task-set}).
  \item We propose a feature-level evaluation for library construction, in which the tests of the original repository grade a delivery feature by feature and the Feature Acceptance Rate (FAR) reports how much of the library works (\Cref{sec:evaluation}).
  \item We evaluate four frontier systems and find that they rarely deliver a fully working library, and we further analyze how they work and how their results change with repository age (\Cref{sec:experiments}).
\end{enumerate}

\section{Related Work}
\label{sec:related}

\begin{table}[t]
  \centering
  \caption{Repository-level coding benchmarks. \yes: the agent writes all the specified code with none of it given, the requirements come as one specification per feature, the repository's own tests grade the result, and the main score counts features. \half: partly so, and \Cref{app:related} gives the detail for every benchmark. Lines per task are changed lines for editing benchmarks and target lines otherwise, as the mean or median each benchmark reports. \nr: not reported.}
  \label{tab:related}
  \small
  \setlength{\tabcolsep}{5pt}
  \renewcommand{\arraystretch}{1.18}
  \begin{tabular}{@{}l>{\centering\arraybackslash}p{1.25cm}>{\centering\arraybackslash}p{1.35cm}>{\centering\arraybackslash}p{1.25cm}>{\centering\arraybackslash}p{1.25cm}rrr@{}}
    \toprule
    Benchmark & From scratch & Spec per feature & Original tests & Score by feature & Lines per task & Tasks & Lang. \\
    \midrule
    SWE-bench & \no & \no & \yes & \no & 33 & 2,294 & 1 \\
    SWE-Bench Pro & \no & \no & \yes & \no & 107 & 1,865 & 4 \\
    FeatureBench & \half & \no & \yes & \no & 790 & 200 & 1 \\
    SWE-Milestone & \no & \half & \yes & \half & 570 & 98 & 5 \\
    \midrule
    Commit0 & \half & \no & \yes & \no & \nr & 54 & 1 \\
    LEGO-REPO & \half & \no & \yes & \no & \nr & 522 & 1 \\
    NL2Repo-Bench & \yes & \no & \yes & \no & 1.5K to 4K & 104 & 1 \\
    BeyondSWE-Doc2Repo & \yes & \no & \half & \no & 3.5K & 50 & 1 \\
    Zero2Repo & \yes & \no & \half & \no & \nr & 17 & 6 \\
    ICAE-Bench & \yes & \no & \half & \no & 35.5K & 480 & 12 \\
    MirrorCode & \yes & \no & \half & \no & 5.9K & 25 & 6 \\
    E2E-SWE & \yes & \no & \no & \no & 7.4K & 186 & 11 \\
    ProgramBench & \yes & \no & \half & \no & 8.6K & 200 & any \\
    PolyCodeEval L0 & \yes & \no & \nr & \no & 2.5K & 58 & 5 \\
    E2EDev & \yes & \half & \no & \half & \nr & 46 & web \\
    RepoCraft & \yes & \no & \half & \no & 97.8K & 6 & 1 \\
    \midrule
    \rowcolor{twbg}\textbf{NL2Repo-Pro} & \yes & \yes & \yes & \yes & \textbf{\pk{med-src-lines-k}K} & \textbf{\pk{tasks-final}} & \textbf{5} \\
    \bottomrule
  \end{tabular}
\end{table}

\paragraph{From functions to long-horizon software work.} Early execution-based benchmarks for code generation graded single functions with unit tests \citep{chen2021evaluatinglargelanguagemodels,austin2021programsynthesislargelanguage}. SWE-bench moved this kind of evaluation into real repositories, where a system resolves an issue with a patch \citep{jimenez2024swebenchlanguagemodelsresolve}. Since then, the reference solutions of such tasks have grown from tens of lines in SWE-bench to over a hundred in SWE-Bench Pro \citep{deng2025swebenchproaiagents} and several hundred in FeatureBench \citep{zhou2026featurebenchbenchmarkingagenticcoding}, which asks for whole features to be implemented inside an existing repository or from scratch. As tasks have grown, scoring has become finer. For example, the long-horizon editing benchmarks SWE-EVO, RoadmapBench and LoLBench report partial credit next to the resolve rate \citep{le2025sweevo,xu2026roadmapbench,peng2026lolbench}. WEAVE likewise gives partial credit, along a dependency graph of pull requests \citep{li2026weave}, and PaperBench grades paper replications with a rubric tree scored by a language model \citep{starace2025paperbenchevaluatingaisability}.

\paragraph{Building a repository from scratch.} A second line of benchmarks asks the agent to write a repository rather than to change one (\Cref{tab:related}). Some of them start from partial code: Commit0 asks for a library to be completed from stubbed functions with its tests visible \citep{zhao2025commit0}, and LEGO-REPO keeps the interfaces and tests of a repository and asks for the implementation \citep{jin2026legorepo}. Others start from an empty workspace, as NL2Repo-Bench does when it asks for a Python library written from one requirements document \citep{ding2026nl2repo}. Among the benchmarks where the agent writes all of the code, what it is given in place of code varies. BeyondSWE-Doc2Repo and E2E-SWE give one specification \citep{chen2026beyondswe,ding2026e2eswe}, Zero2Repo adds an interface contract to a requirements document \citep{yang2026zero2repo}, and ProgramBench and MirrorCode give the original program as an executable \citep{yang2026programbench,adamczewski2026mirrorcode}. ICAE-Bench starts from a short product request and lets the agent question a simulated user \citep{peng2026icaebench}, while RepoCraft and RepoGenesis work from a project description or a README \citep{luo2026rpg,peng2026repogenesis}. The grading varies as well, from the repository's own tests, as in Commit0 and NL2Repo-Bench, to adapted versions of them, as in BeyondSWE-Doc2Repo and Zero2Repo, and to tests written for the benchmark, as in E2E-SWE. Language coverage has widened from one language to as many as twelve, with PolyCodeEval at five and ReCode-bench at seven \citep{yang2026polycodeeval,liu2026recodebench}. In size, however, most benchmarks that report it still target fewer than ten thousand lines per task.

\paragraph{Position of NL2Repo-Pro.} \Cref{tab:related} helps position NL2Repo-Pro among these benchmarks. Of those closest to ours, E2EDev also accepts a requirement only when all of its tests pass, but it covers 46 small web projects with generated tests \citep{liu2025e2edev}. SWE-Milestone gives one document per milestone with a dependency graph and scores each milestone with original tests, but it starts from the code at an earlier release of the repository \citep{deng2026swemilestone}. DeNovoSWE writes the document of each task capability by capability, but it merges the parts into one document per repository and scores runs by their test pass rate \citep{zhao2026denovoswe}. Another line scores generated repositories with a language-model agent instead of a fixed test suite \citep{wang2026verificationhorizon}. None of these benchmarks scores the features of a library built from scratch with the library's own tests. NL2Repo-Pro does so by making the feature the unit of both the specification and the score. Each feature has its own specification, and the specifications are ordered by dependency and given at once. Original tests are not written per feature, so each test is attributed to the features whose code it calls directly. A single delivery can thus be read feature by feature and layer by layer, although the original libraries have a median of \pk{med-src-lines-k} thousand lines of code. NL2Repo-Pro also addresses a problem noted in recent work, that original tests can depend on names a specification leaves out \citep{ding2026e2eswe,ding2026specgaps}: its interface comes from the code the tests call directly, and each task lists the other names they use (\Cref{sec:task-form,sec:construction}).

\section{Designing and Building NL2Repo-Pro}
\label{sec:benchmark}

NL2Repo-Pro consists of \pk{tasks-final} tasks, each built from a real repository in Python, Go, Rust, Java or JavaScript. In each task, an agent receives the requirements of a library, builds it from scratch, and has its delivery graded by the tests of the original repository. Because a library of this size has many features, both the requirements and the score are organized by feature. This section first presents this design, defining a task (\Cref{sec:task-form}) and how it is scored (\Cref{sec:evaluation}). It then describes how the tasks are built, with agents and programs producing each task and engineers reviewing it (\Cref{sec:construction}), and reports the statistics of the task set (\Cref{sec:task-set}).

\subsection{Task Formulation}
\label{sec:task-form}

A task consists of four parts (\Cref{fig:task-form}). Two of them are given to the agent: a requirements specification, which describes the library to build, and an environment, a container with the library's dependencies installed and no network access. The other two stay hidden: the tests of the original repository, which grade the delivery (\Cref{sec:evaluation}), and the original library code, which serves as the reference solution (\Cref{sec:construction}). The agent writes the library from scratch in the environment and delivers it as an installable package.

\paragraph{The requirements specification.} The specification has two levels (\Cref{fig:task-form}, panel 1): an overview document, \texttt{instruction.md}, which describes the library as a whole, and one specification per feature, which states what that feature does. The overview document is the point of entry. It first states the deliverable and the environment and sets the ground rules that hold across the library. It then lists the features\footnote{Task files label each feature as \texttt{Capability $N$} and each dependency layer as \texttt{Group $N$}.} in dependency layers, with a short paragraph per layer: features are grouped by what they use, as recorded in the feature dependency graph, so each layer builds on the ones before it. It closes with a summary of the public interface. The specification of each listed feature is a file in \texttt{/docs}, named by the number and a short title of the feature. Two supporting files complete the specification. \texttt{unspecified-names.md}, also in \texttt{/docs}, lists the names that the tests import or use but no feature specifies, such as error classes and names exposed at the top of a package, each with the path where the agent defines it. \texttt{placed-files.md}, present when the verifier adds original files at grading, gives the paths of these files but not their contents. Most of them are tests, data and configuration.

\paragraph{Feature specifications.} Every feature specification follows the same structure: the context of the feature, its functional requirements, its part of the interface and its acceptance criteria. The context lists the features to read first, since the feature builds on them, and gives its background and scope. The functional requirements then define what the feature does, and the interface fixes how it is called. This interface is written as code stubs: the signatures are binding, and the bodies are left to the agent. The acceptance criteria make the requirements checkable. Each criterion is a concrete scenario that the agent can check against its own code, and it lists the requirements it covers. For example, in the task built from the SQL parser \texttt{sqlglot} (\Cref{app:sample}), the specification of a feature that simplifies SQL expressions and folds constants states \pk{sample-fr} functional requirements, fixes the signature of the function \texttt{simplify} and lists \pk{sample-ac} acceptance criteria. One of these criteria considers a query that contains the sum \mbox{\texttt{7 + 8}}: after the sum is simplified, the query contains \texttt{15} in its place, and a second simplification makes no further change.

\paragraph{Building the library.} The agent starts in an empty workspace,\footnote{In \pk{start-tasks} tasks, \pk{start-rust} of them in Rust, the workspace instead holds starting files, original files that the build or the tests need but no feature specifies (\Cref{app:per-task}).} with the library's dependencies preinstalled at pinned versions and without network access. All documents are available from the start, so the agent decides for itself what to read and what to build first, within a limit of ten hours. The ground rules ask it to use subagents, so that it can divide the library among them. Along the way, the agent never sees the tests, so it can check its work only against the documents, above all the acceptance criteria. The delivery is graded once, after the run ends. \Cref{app:sample} shows a sample task, and \Cref{app:release} describes the released files.

\subsection{Feature-Level Evaluation}
\label{sec:evaluation}

In this subsection, we describe how NL2Repo-Pro evaluates a delivery. The guiding principle is to score a delivery by its features rather than by its tests, so that the score shows how much of the library works. We first explain how the original tests grade a delivery and how each test is attributed to the features it calls. We then define the main score, the Feature Acceptance Rate (FAR), on these per-feature results, and finally introduce two secondary scores that relate our results to earlier benchmarks.

\paragraph{Grading and attribution.} Evaluating a delivery has two parts: grading, which determines the tests the delivery passes, and attribution, which determines the features each test counts for. For grading, the verifier, which uses no language model, installs the delivered package in a separate container, restores the original repository's tests and adds the listed original files. It then runs the graded tests of the task, a list fixed when the task is released. A graded test that does not run, for instance because the delivery fails to compile or import, counts as failed, so every delivery is graded on the same list. For attribution, each graded test counts for every feature whose code it calls directly (\Cref{app:attribution}). A call is direct when it is made by test code, including fixtures and test helpers, and not by the library itself.

\paragraph{Feature Acceptance Rate.} A feature is accepted when every graded test attributed to it passes, and the FAR of a task is the share of its features that are accepted (\Cref{fig:task-form}). Formally, let $G_f$ be the graded tests attributed to feature $f$, and let $F$ be the features of a task with $G_f \neq \emptyset$. The FAR of the task is
\begin{equation}
  \mathrm{FAR} = \frac{1}{|F|} \sum_{f \in F} \mathbf{1}\bigl[\,\text{every test in } G_f \text{ passes}\,\bigr].
  \label{eq:far}
\end{equation}
The rule is strict by design, as in acceptance testing, where a feature ships only when all of its acceptance tests pass. A feature with more tests must pass more cases to be accepted. It still counts as one feature, whereas in a test pass rate its weight grows with the number of its tests (\Cref{sec:task-set}). \Cref{app:attribution} measures how acceptance changes with the number of tests.

We report FAR averaged over tasks. The features of a task are part of its release (\Cref{app:release}), so every system is graded on the same features. Additionally, the evaluation offers more than a single score per task, because every feature has its own verdict. These verdicts can be read layer by layer, since the features are ordered by dependency (\Cref{sec:main-results}).

\paragraph{Secondary scores.} Besides FAR, we report the two usual scores, which relate our results to earlier benchmarks. The first is the test pass rate, the share of a task's graded tests that pass, averaged over tasks, and it is the main score of NL2Repo-Bench \citep{ding2026nl2repo}. The second is the number of tasks with every feature accepted. This is a whole-task verdict, and it corresponds to the resolve rate of issue-level benchmarks such as SWE-bench \citep{jimenez2024swebenchlanguagemodelsresolve}.

\begin{figure}[t]
  \centering
  \includegraphics[width=\linewidth]{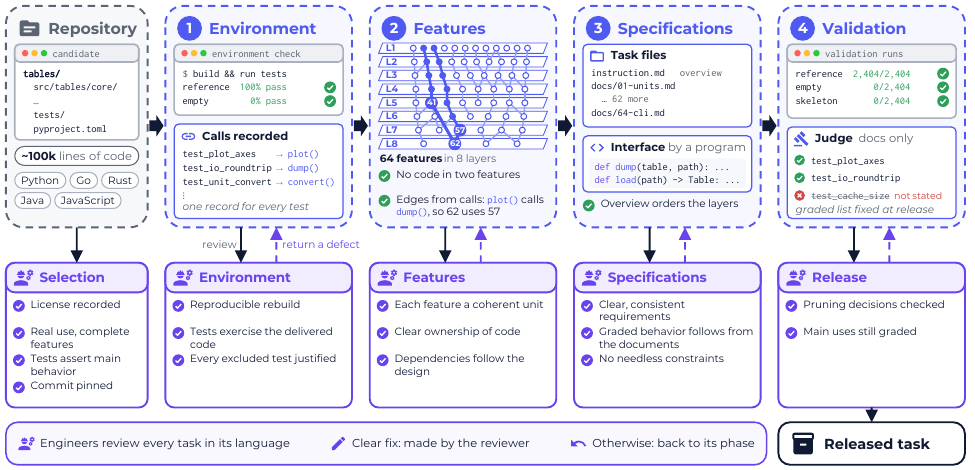}
  \caption{Construction of a task. Agents and programs produce each task in four phases (top),
  and each phase shows what it produces and guarantees. Engineers review every task at five
  gates (bottom), and \Cref{tab:gates} lists what each gate secures. Solid arrows pass work to review, and
  dashed arrows return a defect to the phase that produced it. All artifacts are
  illustrative.}
  \label{fig:pipeline}
\end{figure}

\subsection{Construction}
\label{sec:construction}

We select repositories whose library is in real use and has tests for its main behavior. Among them, we keep only those that can be built and tested offline on ordinary hardware. Construction then turns each of these repositories into a task and checks it at both extremes. The reference solution, which is the original library code, passes every graded test, so the task is solvable. When we leave its function bodies unimplemented, it scores zero. Both checks can be rerun from the released files.

Each selected repository passes through four phases, in which agents and programs produce the material of the task and engineers review it at the end of each phase (\Cref{fig:pipeline}). The first phase builds an executable environment, the second divides the library into features and computes their dependencies, the third writes the specification, and the fourth packages and validates the task. A repository is dropped when it cannot yield a complete task, for instance when its library cannot be built offline or its features do not depend on one another. We summarize the phases and the review below, and \Cref{app:construction} gives the full account.

\paragraph{Four phases.} Task construction is an automated pipeline built on agents: agents do the open-ended work of each phase, and programs compute what follows directly from the code. First, an agent builds an environment in which the original code passes every retained test and an empty workspace passes none. This phase also records which library code each test calls directly. Second, with the environment in place, a planning agent divides the library into features, so that no piece of code belongs to more than one of them. A program then computes the feature dependency graph from the calls between features, and a repository whose features do not depend on one another is dropped. Third, a specification agent writes one specification for each of these features. A program, not the agent, derives the interface section from the code that the tests call directly, as recorded in the first phase, and the overview document orders the specifications along the dependency graph. Fourth, once the documents are complete, the task is packaged with the original code as its reference solution, and validation runs confirm that it passes every graded test and that an empty workspace scores zero. A judge agent then reads each graded test against the documents, without the original code, and removes from grading the tests that it judges to check behavior the documents do not state.

\paragraph{Human review.} Every phase of this automated pipeline is checked by human reviewers. Software engineers who work in the language of a task review it at five gates, one before construction and one at the end of each phase (\Cref{fig:pipeline}, bottom). When a defect has a clear fix, the reviewer corrects it directly. Otherwise, the defect goes back to the phase that produced it, and the task is rebuilt from there.

\subsection{Benchmark Statistics}
\label{sec:task-set}

\begin{table}[t]
  \centering
  \caption{Size of the tasks by language, as medians per task. Features counts every feature the specification describes, with or without graded tests. Layers are dependency layers.
  FRs and ACs are functional requirements and acceptance criteria. Interface counts the
  declarations in the interface sections, and tokens cover all documents the agent reads.
  Lines count the non-blank source lines of the original library in its language, outside test directories, for reference.}
  \label{tab:stats}
  \small
  \begin{tabular}{@{}lrrrrrrrrr@{}}
\toprule
Language & Tasks & Features & Layers & FRs & ACs & Interface & Tokens & Graded tests & Lines \\
\midrule
\langlogo{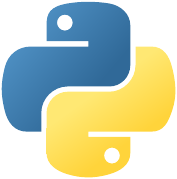}~Python & 70 & 50 & 9 & 1,080 & 493 & 681 & 180K & 891 & 75K \\
\langlogo{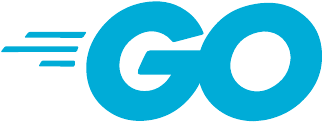}~Go & 20 & 151 & 18 & 3,118 & 1,579 & 3,330 & 577K & 3,579 & 131K \\
\langlogo{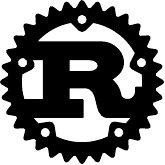}~Rust & 20 & 37 & 8 & 984 & 442 & 835 & 164K & 683 & 105K \\
\langlogo{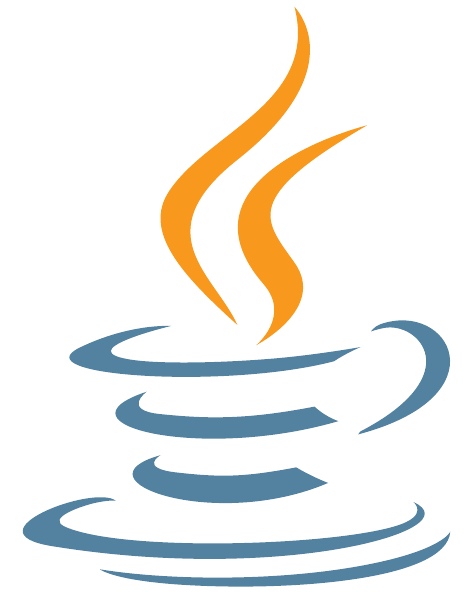}~Java & 20 & 41 & 12 & 882 & 344 & 1,011 & 124K & 281 & 39K \\
\langlogo{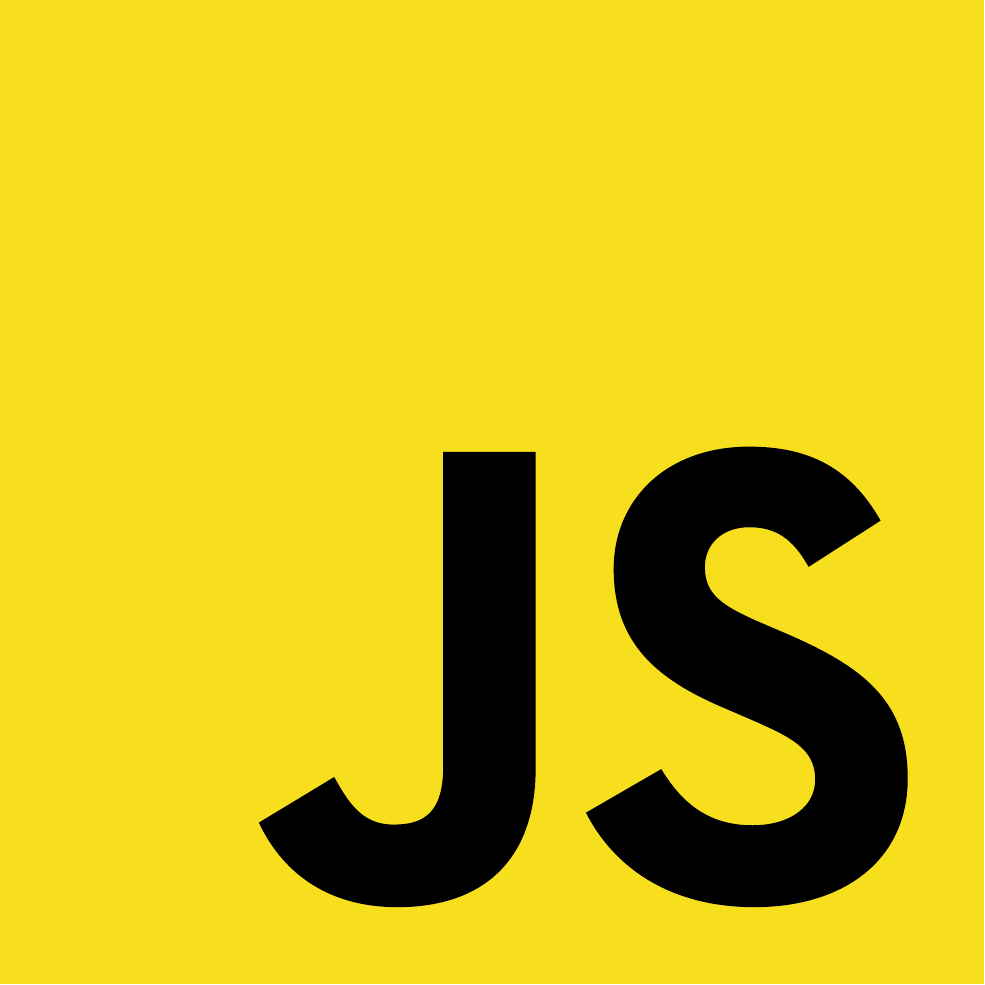}~JavaScript & 20 & 46 & 5 & 533 & 314 & 146 & 97K & 214 & 30K \\
\midrule
\langspace~All & 150 & 51 & 9 & 1,042 & 490 & 795 & 180K & 750 & 74K \\
\bottomrule
\end{tabular}

\end{table}

This subsection reports the statistics of the \pk{tasks-final} tasks. It moves from the task set as a whole to the inside of a task: the repositories behind the tasks, the specification that the agent reads, the tests that grade a delivery, and the features labeled as core.

\begin{figure}[t]
  \centering
  \includegraphics[width=\linewidth]{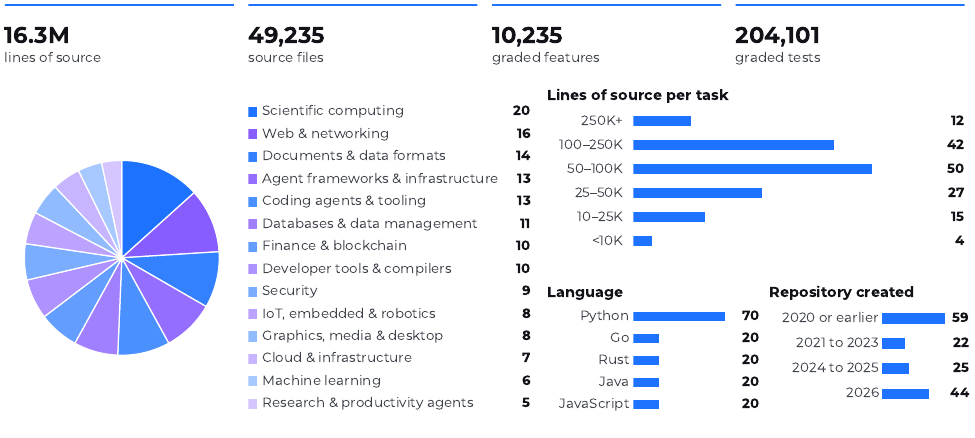}
  \caption{The task set. Totals across all \pk{tasks} tasks, then the tasks by project type, size of the original library, language and the year the repository was created.}
  \label{fig:composition}
\end{figure}

\paragraph{Repositories.} The set spans \pk{types} project types and five languages. Scientific computing contributes the most tasks, followed by web and networking (\Cref{fig:composition}). The original libraries have a median of \pk{med-src-lines-k} thousand lines of code, with an interquartile range from \pk{q1-src-lines-k} to \pk{q3-src-lines-k} thousand. The pinned commits are also recent: of the \pk{tasks} tasks, \pk{commits-2026} are pinned to a commit from 2026, and \pk{created-2026} repositories were created in 2026. \Cref{sec:recall} compares results on these repositories with those on older ones. \Cref{tab:tasks} lists every task with its language, license class, size and results.

\paragraph{Specifications.} The specification of a task covers a median of \pk{med-spec-features} features in \pk{med-groups} dependency layers, with \pk{min-groups} to \pk{max-groups} layers per task (\Cref{tab:stats}). Its feature specifications state a median of \pk{med-fr} functional requirements and \pk{med-ac} acceptance criteria. Together, the documents of a task total a median of about \pk{med-tokens-k} thousand tokens, and the largest task reaches \pk{max-tokens-m} million.

\paragraph{Graded tests.} A task grades a median of \pk{med-graded-tests} tests. Of the \pk{features-all} features in the set, \pk{features-graded} carry graded tests, a median of \pk{med-features} per task. These tests are spread unevenly over the features. Within a task, the median feature has \pk{med-tests-per-feature} graded tests, as a median over tasks. The most heavily tested feature, by contrast, carries a median of \pk{med-top-feature-share}\% of the task's graded tests, and in one task a single feature carries \pk{max-feature-tests}. This unevenness is why FAR counts features rather than tests (\Cref{sec:evaluation}).

\paragraph{Core features.} During construction, software engineers labeled every feature by hand as core or not. A feature is core when its main work is substantial implementation, such as algorithms, protocols, file formats, numerical methods, concurrency or invariants that span modules. Peripheral work such as configuration, validation, adapters, command-line entry points and error handling is not core, however important it is to users. Of the graded features, \pk{core-graded} (\pk{core-share}\%) are core, and \Cref{sec:main-results} compares the systems on them.

\section{Experiments}
\label{sec:experiments}

In this section, we evaluate four frontier systems on NL2Repo-Pro. After describing the setup (\Cref{sec:setup}), we present the main results, which show how much of a library each system delivers and what FAR reveals beyond the usual scores (\Cref{sec:main-results}). We then analyze how the systems work through the layered specification (\Cref{sec:work}) and how their results change with repository age (\Cref{sec:recall}).

\subsection{Setup}
\label{sec:setup}

We evaluate four systems, each a model paired with an agent harness, so each result describes a model together with its harness. GPT-6.1 Sol runs in Codex CLI, and Claude Opus 5.5, DeepSeek V4.1 Flash and Grok 4.7 run in Claude Code. Every model runs at the xhigh reasoning effort, and \Cref{app:config} lists the settings of every system. All systems run under the same conditions: each task has 4 CPU cores, 32 GB of memory and 64 GB of storage, and the agent has ten hours and no network access. The ground rules of every task also ask the agent to use subagents, so the harnesses work under the same rule as well. Once the agent has finished, the verifier grades the delivery in a separate container, within six and a half hours. A run is a single execution of a system on a task, from the initial workspace to the graded delivery, and the main results use one run per system and task, \pk{runs-lead-total} runs in all. We report 95\% bootstrap intervals over tasks, computed with 10,000 resamples, and measure resource use in tokens, subagents and tool calls.

\subsection{Main Results}
\label{sec:main-results}

\begin{table}[t]
  \centering
  \caption{Main results, with one run per system and task. FAR is the main score, with its 95\% bootstrap interval over tasks.
  All features counts the tasks with every feature accepted, which here is the same as every
  graded test passing. The language columns give FAR. Input and output are token counts,
  medians per run over the main agent and all subagents, and input is cumulative (\Cref{app:config}).}
  \label{tab:main}
  \scriptsize
  \setlength{\tabcolsep}{3pt}
  \begin{tabular}{@{}lrcrrrrrrrrr@{}}
\toprule
System & FAR & 95\% interval & Test pass rate & All features & Python & Go & Rust & Java & JavaScript & Input & Output \\
\midrule
\syslogo{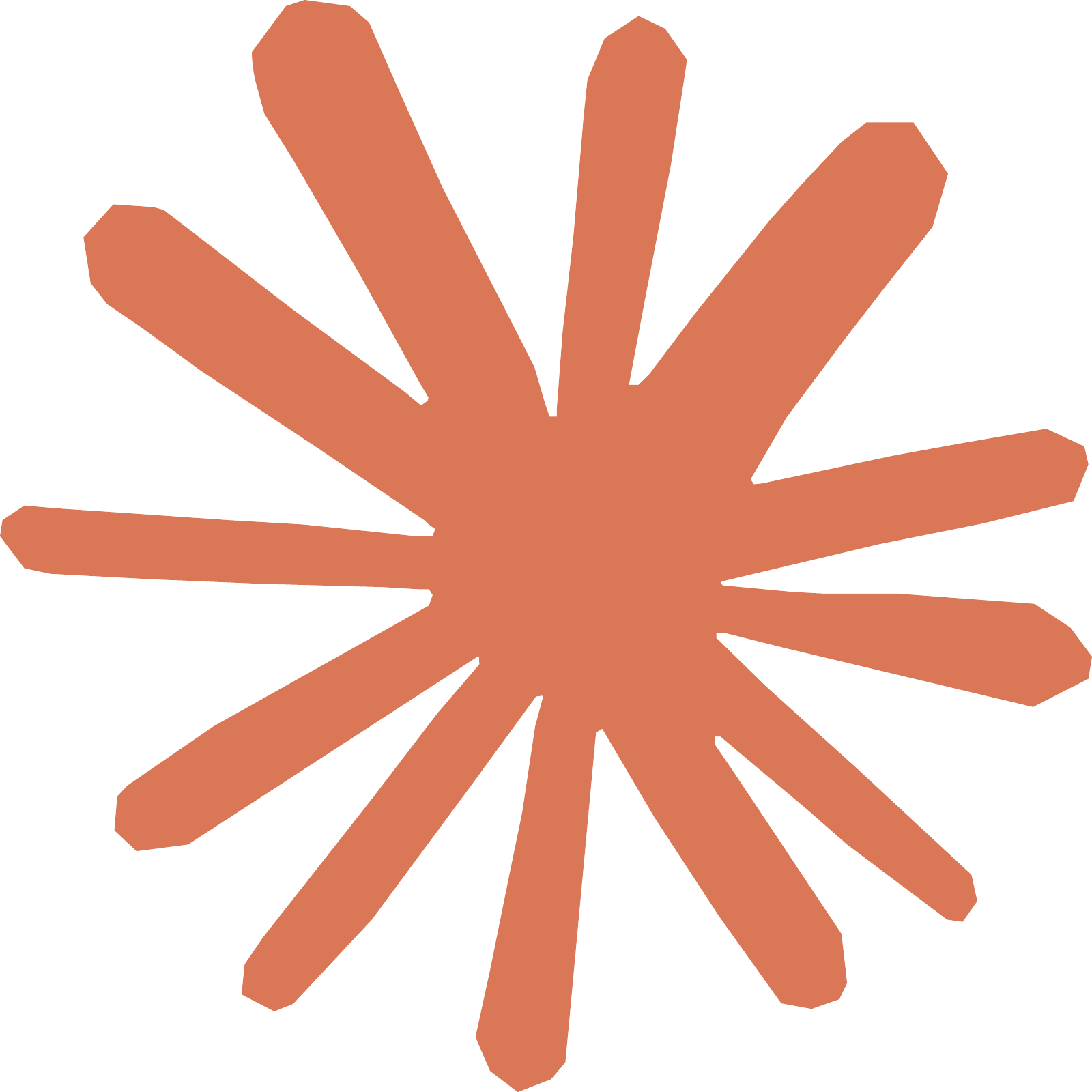}~Claude Opus 5.5 & \textbf{61.7} & [57.5, 65.9] & 87.2 & 7 & 58.9 & 57.6 & 59.2 & 61.0 & 78.6 & 192.1M & 2.8M \\
\syslogo{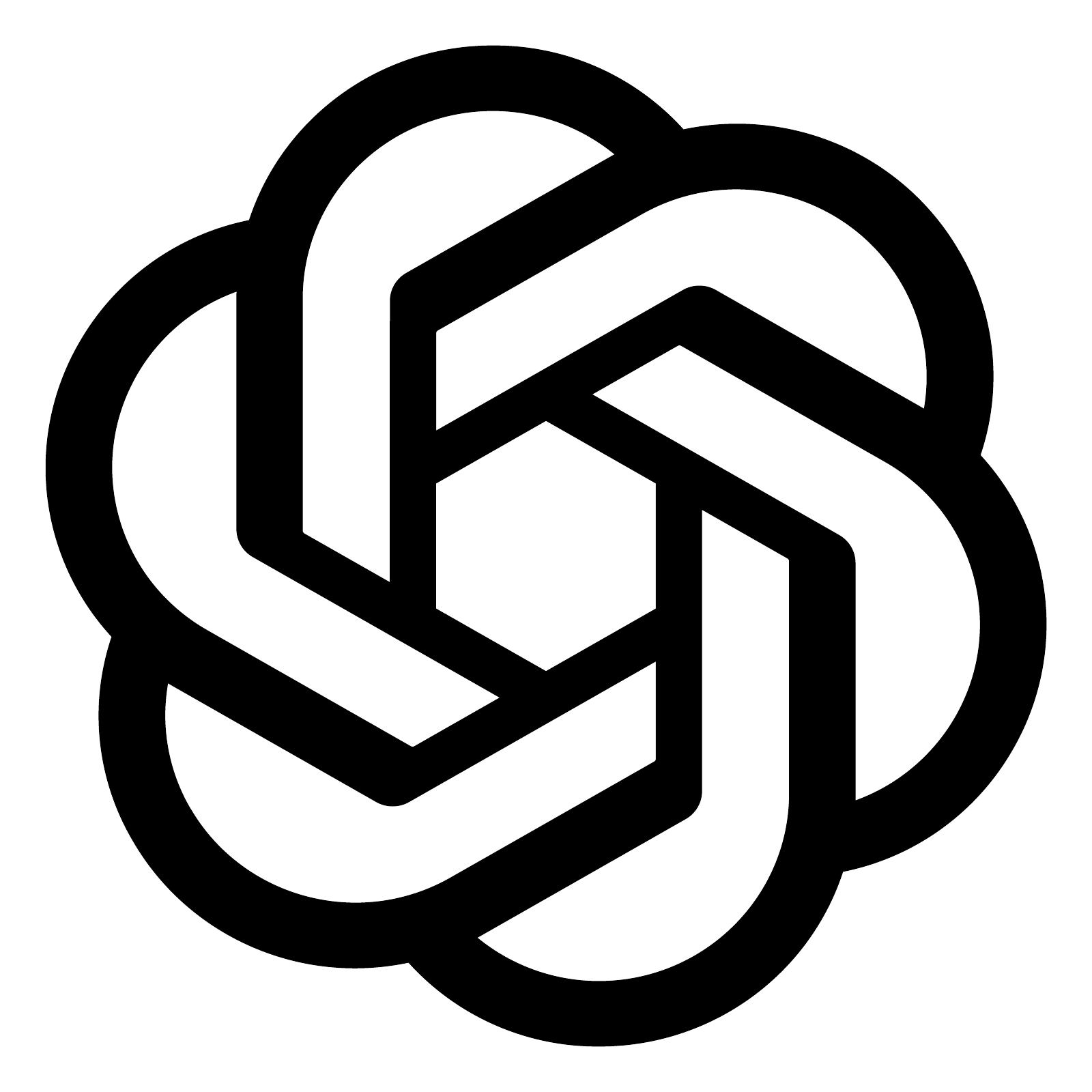}~GPT-6.1 Sol & \textbf{47.1} & [43.6, 50.5] & 81.8 & 0 & 39.6 & 49.7 & 49.9 & 43.2 & 71.7 & 65.1M & 308K \\
\syslogo{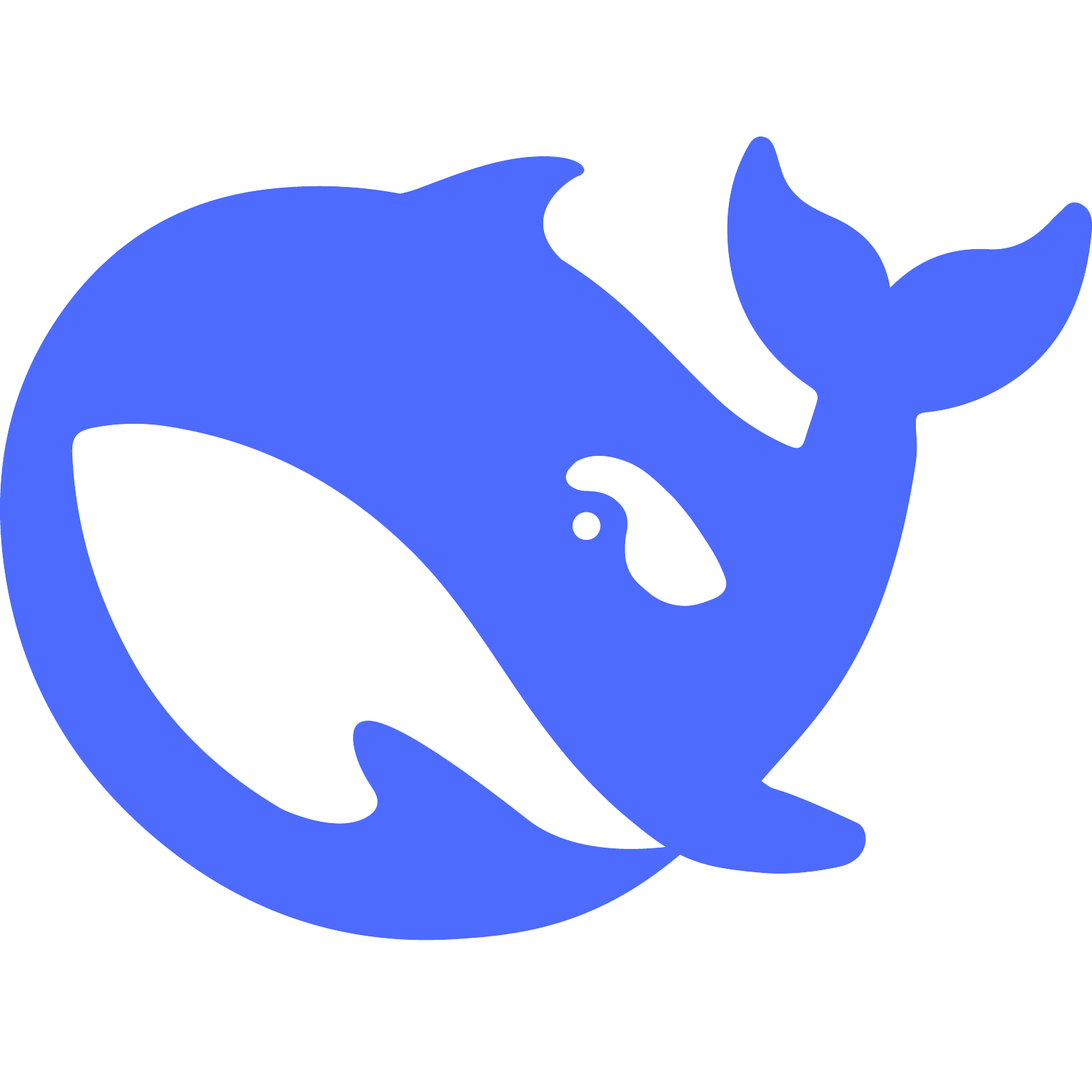}~DeepSeek V4.1 Flash & \textbf{37.7} & [34.4, 41.1] & 73.5 & 0 & 31.5 & 43.4 & 35.0 & 32.9 & 61.5 & 518.7M & 2.1M \\
\syslogo{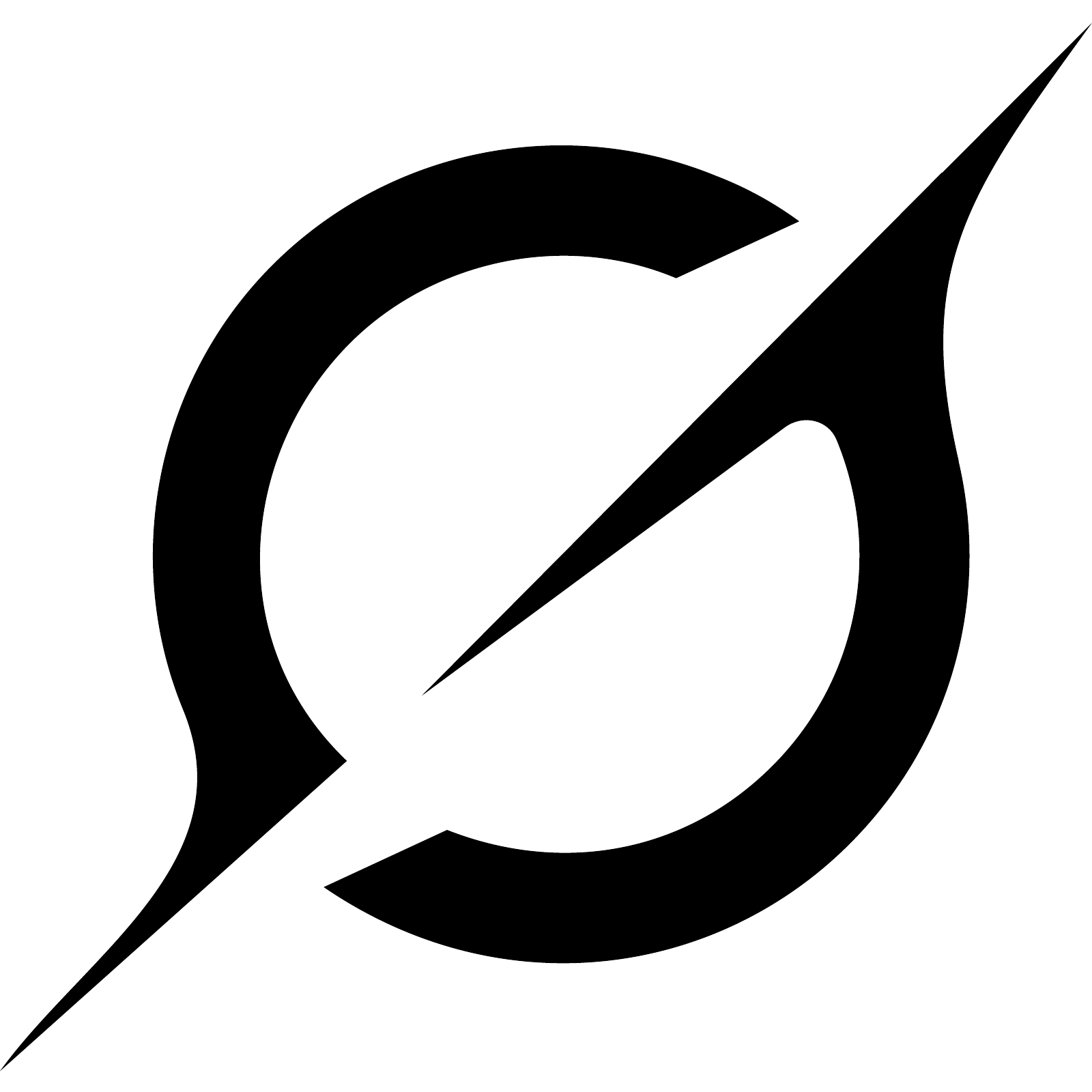}~Grok 4.7 & \textbf{22.5} & [19.4, 25.7] & 51.4 & 0 & 15.5 & 25.1 & 18.5 & 23.3 & 47.4 & 85.1M & 634K \\
\bottomrule
\end{tabular}

\end{table}

\paragraph{Overall results.} \finding{Frontier systems rarely deliver a fully working library.} Of the \pk{runs-lead-total} runs in \Cref{tab:main}, only \pk{resolve-total} have every feature accepted, and all of them are runs of Claude Opus 5.5. How much of a library each system delivers is measured by FAR, our main score. Claude Opus 5.5 ranks first at \pk{opus-far}\%, followed by GPT-6.1 Sol at \pk{sol-far}\%, DeepSeek V4.1 Flash at \pk{deepseek-far}\% and Grok 4.7 at \pk{grok-far}\%. The lead of Claude Opus 5.5 over GPT-6.1 Sol is \pk{opus-delta-abs} points (95\% interval \pk{opus-delta-lo} to \pk{opus-delta-hi}), and it scores higher on \pk{opus-wins} of the \pk{opus-common} tasks.

\paragraph{Results across languages.} This ranking is the same in every language: the four systems rank in the same order in all five (\Cref{tab:main}). The order also holds in \pk{types-same-order} of the \pk{types} project types and in every age group of repositories, including the \pk{new-n} created in 2026 (\Cref{sec:recall}), and it holds on core features as well (\Cref{sec:task-set}). Counted over the core features of each task that has them, FAR is \pk{opus-corefar}\% for Claude Opus 5.5, \pk{sol-corefar}\% for GPT-6.1 Sol, \pk{deepseek-corefar}\% for DeepSeek V4.1 Flash and \pk{grok-corefar}\% for Grok 4.7. The size of the lead, however, differs between languages. Claude Opus 5.5 leads GPT-6.1 Sol by \pk{lead-python} points on Python and \pk{lead-java} on Java, against \pk{lead-rest-min} to \pk{lead-rest-max} on Go, Rust and JavaScript. The lead is smallest on JavaScript, where every system reaches its highest score and the tasks have the fewest layers and the shortest documents (\Cref{tab:stats}).

\begin{figure}[t]
  \centering
  \includegraphics[width=\linewidth]{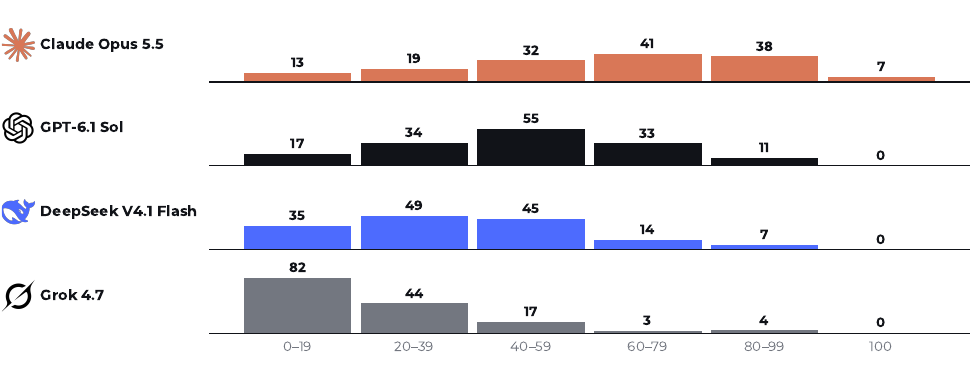}
  \caption{Number of tasks in each FAR band, with one run per system and task. The rightmost band holds the tasks with every feature accepted, the only tasks that a whole-task verdict credits.}
  \label{fig:bands}
\end{figure}

\paragraph{FAR and the usual scores.} The two usual scores each miss how much of a library works, in opposite directions. The whole-task verdict credits only the rightmost band of \Cref{fig:bands}, which is empty for three of the four systems, so it scores almost every run as a failure. The test pass rate, in contrast, overstates how much works. Claude Opus 5.5 passes \pk{opus-t}\% of the graded tests, which suggests a nearly finished library, yet on average more than a third of its features in a task fail at least one test, and its FAR is \pk{opus-far}\%. For every system, the test pass rate exceeds FAR by \pk{gap-min} to \pk{gap-max} points (\Cref{tab:main}). The test pass rate also blurs the gap between strong systems: on the \pk{hi95-n} tasks where Claude Opus 5.5 and GPT-6.1 Sol both pass at least 95\% of the graded tests, their test pass rates differ by \pk{hi95-t-gap} points and their FAR by \pk{hi95-far-gap}.

\paragraph{Failure analysis.} Since every feature has its own verdict, we can also locate where libraries fail. Many failures are shared by all systems: of all graded features, \pk{none-share}\% are accepted for none of the four. Failures also vary with where a feature sits in the dependency order. For every system, features in the first third of a task's layers are accepted more often than those in the middle or last third. Within a library, the fitted share of features accepted falls from the first layer to the last by \pk{opus-depth-slope} points for Claude Opus 5.5, \pk{sol-depth-slope} for GPT-6.1 Sol, \pk{deepseek-depth-slope} for DeepSeek V4.1 Flash and \pk{grok-depth-slope} for Grok 4.7 (\Cref{app:attribution}). The stronger the system, the smaller this fall.

\begin{figure}[t]
  \centering
  \includegraphics[width=\linewidth]{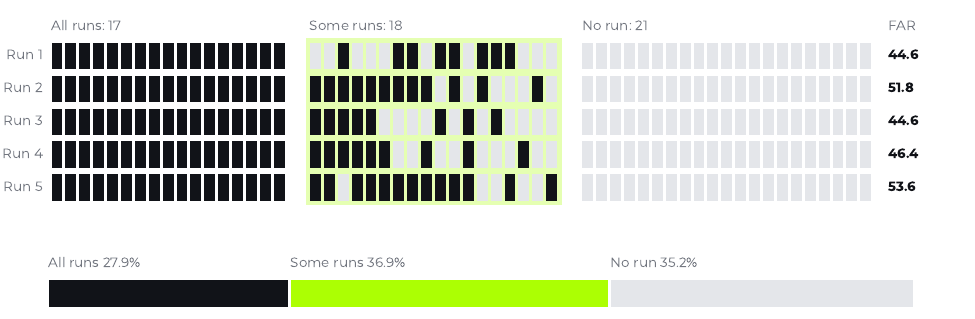}
  \caption{Five runs of GPT-6.1 Sol. Top: each column is one feature of \texttt{\pk{grid-name}}, filled in the runs where it is accepted, and the highlighted features change from run to run. Bottom: the share of features over all \pk{tasks} tasks.}
  \label{fig:runs}
\end{figure}

\paragraph{Variation between runs.} To see how much a result depends on a single run, we run GPT-6.1 Sol five times on each task, and the main results use its first run. Between runs, the mean score barely moves while single features change. The mean FAR of each of the five runs over all tasks stays between \pk{sol5-run-min}\% and \pk{sol5-run-max}\%, yet \pk{sol5-sometimes}\% of all features are accepted in some runs and rejected in others, against \pk{sol5-always}\% accepted in every run and \pk{sol5-never}\% in none (\Cref{fig:runs}). This variation suggests room for test-time scaling: the best of five runs reaches a FAR of \pk{sol5-best5}\%, against \pk{sol5-best1}\% for a single run on average.

\subsection{Behavior Analysis}
\label{sec:work}

\begin{table}[t]
  \centering
  \caption{How the four systems work, medians per run unless marked. Input and output tokens cover the main agent and all subagents, and input is cumulative (\Cref{app:config}).}
  \label{tab:work}
  \footnotesize
  \setlength{\tabcolsep}{4pt}
  \begin{tabular}{@{}lrrrr@{}}
\toprule
 & \syslogo{claude}~Claude Opus 5.5 & \syslogo{gpt}~GPT-6.1 Sol & \syslogo{deepseek}~DeepSeek V4.1 Flash & \syslogo{grok}~Grok 4.7 \\
\midrule
Cumulative input tokens & 192.1M & 65.1M & 518.7M & 85.1M \\
Output tokens & 2.8M & 308K & 2.1M & 634K \\
Context peak, main agent & 231K & 271K & 521K & 248K \\
Subagents spawned & 25 & 3 & 13 & 8 \\
Tool calls & 1,371 & 399 & 2,524 & 981 \\
Runs with a shared plan file & 93\% & 4\% & 38\% & 53\% \\
Runs where the main agent writes code & 14\% & 100\% & 87\% & 88\% \\
Test commands & 130 & 8 & 10 & 0 \\
Source lines written & 50.4K & 13.7K & 46.1K & 26.9K \\
\bottomrule
\end{tabular}

\end{table}

Scores show how much of a library a system delivers, but not how it works. In this subsection, we analyze four aspects of the systems' behavior: how many resources they use, how they organize their subagents, how they check their work, and how they plan it. We draw on two sources. The first is a set of measures computed from the trajectory of every run (\Cref{tab:work}). The second is a set of labels: software engineers mark every run in the main results for 23 behaviors and quote the trajectory for each behavior they mark present. We report each behavior as a share of the runs in which it could be judged (\Cref{app:config}).

\paragraph{Resource use.} The systems differ in resource use by nearly an order of magnitude (\Cref{tab:work}). Claude Opus 5.5 uses a median of \pk{opus-w-out} output tokens per run, and GPT-6.1 Sol \pk{sol-w-out}. Time is rarely what stops a run: nearly every run ends on its own, and only \pk{hit10h-total} of the \pk{runs-lead-total} runs reach the ten-hour limit, \pk{opus-hit10h} of Claude Opus 5.5 and \pk{deepseek-hit10h} of DeepSeek V4.1 Flash. Within one system, using more resources need not go with a higher score: among the five runs of GPT-6.1 Sol on the same task, the runs with more output tokens score no higher, apart from a few runs that score far below the others on the same task, most of which did not build or load completely (\Cref{app:config}).

\paragraph{Use of subagents.} The ground rules ask every system to use subagents, but the systems organize them differently. At one end, Claude Opus 5.5 works as a coordinator: it spawns a median of \pk{opus-w-subagents} subagents, writes a shared plan file in most runs, and its main agent rarely writes code. At the other, GPT-6.1 Sol works alongside its subagents: it spawns \pk{sol-w-subagents}, as many as its harness runs at once, and its main agent writes code in every run. DeepSeek V4.1 Flash and Grok 4.7 lie between the two (\Cref{tab:work}).

\paragraph{Checking.} With the tests hidden, every system checks its work against the specification. In \pk{b-min-spec-example-checks}\% to \pk{b-max-spec-example-checks}\% of runs, depending on the system, the agent compares its code with the acceptance criteria and other concrete examples in the specification, and in \pk{b-min-interface-audit}\% to \pk{b-max-interface-audit}\% a script also checks that every name and signature the specification lists is present. The most distinctive behavior is that of Grok 4.7, which mostly stops at import or compile checks. It runs no test command in \pk{grok-no-test-runs}\% of its runs, and in \pk{grok-last-smoke} of its \pk{tasks} runs its last check is an import or compile check, although its runs that do run tests score no higher than the rest. Grok 4.7 is also the system that most often reports more than it delivers. Its closing message overstates the work in \pk{b-grok-completion-overclaim}\% of its runs, against \pk{overclaim-rest-min}\% to \pk{overclaim-rest-max}\% for the other systems, and at grading \pk{grok-not-run}\% of its tests never start, against \pk{deepseek-not-run}\% to \pk{sol-not-run}\%.

\paragraph{Planning.} The task leaves the order of work to the agent, and the systems use the layers as a guide to dependencies rather than as a fixed schedule. Few runs start from the whole specification: only \pk{b-min-full-spec-read-first}\% to \pk{b-max-full-spec-read-first}\% read every feature specification before writing code, and the others open the specifications as the work proceeds, loosely in the order of their numbers (\Cref{app:config}). The systems also do not finish one layer before starting the next. In \pk{ord-started-min}\% to \pk{ord-started-max}\% of adjacent layer pairs, the later layer starts before every feature of the earlier one has a first implementation, and they rarely test one layer before moving on, with a test command between the two layers in at most \pk{ord-gated-max}\% of pairs. Their order still follows the dependencies. In a median run, \pk{ord-edge-min}\% to \pk{ord-edge-max}\% of the dependencies, depending on the system, are built in order, with a feature first implemented before the features that use it. A shuffled order gives \pk{ord-null-min}\% to \pk{ord-null-max}\%.

\subsection{Results by Repository Age}
\label{sec:recall}

\begin{table}[t]
  \centering
  \caption{Results by the year the repository was created. Line overlap is the median share of a
  system's code lines that match the original, after both are formatted with the same standard
  tool. Hidden test names counts the tasks where a run
  writes at least 1\% of the names of the hidden tests. Stated intent to rebuild is the share of runs, among those whose trajectory keeps reasoning
  text, whose reasoning states an intent to rebuild the original from memory. The reasoning of
  GPT-6.1 Sol is encrypted. The reasoning of Claude Opus 5.5 is a summary, so its share is compared only across its own age groups. Logos mark the systems as in \Cref{tab:main}.}
  \label{tab:age}
  \scriptsize
  \setlength{\tabcolsep}{2.4pt}
  \begin{tabular}{@{}lr*{16}{>{\raggedleft\arraybackslash}p{2.1em}}@{}}
\toprule
& & \multicolumn{4}{c}{FAR (\%)} & \multicolumn{4}{c}{Line overlap (\%)} & \multicolumn{4}{c}{Hidden test names} & \multicolumn{4}{c}{Intent to rebuild (\%)} \\
\cmidrule(lr){3-6}\cmidrule(lr){7-10}\cmidrule(lr){11-14}\cmidrule(l){15-18}
Created & Tasks & \syslogo{claude} & \syslogo{gpt} & \syslogo{deepseek} & \syslogo{grok} & \syslogo{claude} & \syslogo{gpt} & \syslogo{deepseek} & \syslogo{grok} & \syslogo{claude} & \syslogo{gpt} & \syslogo{deepseek} & \syslogo{grok} & \syslogo{claude} & \syslogo{gpt} & \syslogo{deepseek} & \syslogo{grok} \\
\midrule
2020 or earlier & 59 & 66.3 & 44.1 & 33.0 & 19.3 & 50.4 & 14.6 & 16.9 & 19.2 & 20 & 0 & 4 & 1 & 82 & -- & 98 & 80 \\
2021 to 2023 & 22 & 65.9 & 46.4 & 40.1 & 21.8 & 30.0 & 11.8 & 16.8 & 19.0 & 7 & 1 & 1 & 1 & 86 & -- & 91 & 59 \\
2024 to 2025 & 25 & 55.7 & 50.3 & 36.3 & 19.9 & 18.1 & 14.8 & 15.1 & 18.8 & 2 & 0 & 1 & 0 & 75 & -- & 80 & 44 \\
2026 & 44 & 56.8 & 49.5 & 43.6 & 28.5 & 15.1 & 14.5 & 15.2 & 18.1 & 0 & 0 & 0 & 0 & 41 & -- & 57 & 23 \\
\bottomrule
\end{tabular}

\end{table}

The tasks come from real repositories, so a model may have seen the original code in training. During a run, the agent has no network access and its workspace holds none of the original code of the specified features, but the model may still remember that code. To see how much this matters, we group the tasks by the year their repository was created (\Cref{tab:age}). The \pk{created-2026} repositories created in 2026 are the least likely to appear in training data (\Cref{sec:task-set}). We first look for signs of recall in each age group and then compare the scores.

\paragraph{Signs of recall.} We look for three signs that a system recalls the original code (\Cref{app:recall}). Hidden test names counts the tasks on which a run writes at least 1\% of the names of the hidden tests, which the task never shows. Line overlap is the share of the code lines of a delivery that match a line of the original, after both are formatted with the same standard tool. Stated intent to rebuild is a statement in the reasoning of a run that it will rebuild the original from memory. These signs fade as repositories get newer, most clearly for Claude Opus 5.5 (\Cref{tab:age}). On the \pk{old-n} repositories from 2020 or earlier, it writes at least 1\% of the hidden test names on \pk{opus-old-names} tasks, against \pk{deepseek-old-names} for DeepSeek V4.1 Flash, \pk{grok-old-names} for Grok 4.7 and none for GPT-6.1 Sol, and its line overlap reaches \pk{opus-old-overlap}\%. On the \pk{new-n} repositories created in 2026, no system writes even 1\% of the hidden test names, and the line overlap of every system lies between \pk{base-min}\% and \pk{base-max}\%. We take this level as a baseline, since lines can also match because they follow the interface in the specification. Only Claude Opus 5.5 rises far above its baseline on older repositories, while the other three stay within a few points of theirs. Stated intent follows the same trend: every system whose reasoning can be read states an intent to rebuild in more runs on the oldest repositories than on the newest.

\paragraph{Scores by repository age.} The lead of Claude Opus 5.5 over GPT-6.1 Sol is also much smaller on newer repositories. It is \pk{lead-old} points on repositories from 2020 or earlier and \pk{lead-y2021} on those from 2021 to 2023, against \pk{lead-y2024} on those from 2024 to 2025 and \pk{lead-new} on those created in 2026 (\Cref{tab:age}). Between the oldest and the newest group, it changes by \pk{lead-change} points (95\% interval \pk{lead-change-lo} to \pk{lead-change-hi}). This change should not be read as an effect of recall alone. The age groups also differ in their tasks, and within a task, the features for which Claude Opus 5.5 writes more of the original code are accepted no more often than the others (\Cref{app:recall}). The lead shrinks from both sides: from the oldest to the newest repositories, the FAR of Claude Opus 5.5 falls, while that of the other three systems rises.

\section{Conclusion}
\label{sec:conclusion}

NL2Repo-Pro asks coding agents to build production-scale libraries from scratch and grades each delivery feature by feature with the tests of the original repository. Beyond the individual findings, our results carry broader implications. As the unit of work grows to a whole library, a single verdict or a single pass rate no longer describes a delivery, so evaluation has to show how much of it works. What a system builds from a long-known repository can differ from what it builds from a new one, so benchmarks at this scale need to keep pace with new code. Most importantly, the frontier of coding agents has shifted: the strongest system already builds most of a library but rarely a complete one, and delivering software that works in full remains an open challenge.

\bibliographystyle{assets/plainnat}
\bibliography{references}

\clearpage
\beginappendix
\crefalias{section}{appendix}
\crefalias{subsection}{subappendix}

\section{Related Benchmarks}
\label{app:related}

This appendix expands \Cref{tab:related}. \Cref{tab:related-full} gives the starting point, requirements, grading and size of each benchmark.

\begin{table}[!h]
  \centering
  \caption{Repository-level coding benchmarks in detail. Start is the code the agent begins with. Score is what the main score counts: the task as a whole-task verdict, the tests as a test pass rate, or the feature. Size is per task, as changed lines for editing benchmarks and as target lines otherwise, with the mean or median as each benchmark reports it, so the measures differ across benchmarks. ProgramBench lets agents write in any language. In \pk{start-tasks} NL2Repo-Pro tasks, the workspace holds starting files, original files that the build or the tests need but that no feature specifies (\Cref{app:per-task}). \nr: not reported.}
  \label{tab:related-full}
  \scriptsize
  \setlength{\tabcolsep}{3pt}
  \renewcommand{\arraystretch}{1.15}
  \begin{tabular}{@{}>{\raggedright\arraybackslash}p{2.3cm}>{\raggedright\arraybackslash}p{1.4cm}>{\raggedright\arraybackslash}p{2.7cm}>{\raggedright\arraybackslash}p{2.1cm}>{\raggedright\arraybackslash}p{1.45cm}>{\raggedright\arraybackslash}p{1.65cm}r>{\raggedright\arraybackslash}p{0.75cm}@{}}
    \toprule
    Benchmark & Start & Requirements & Graded by & Score & Size & Tasks & Lang. \\
    \midrule
    \multicolumn{8}{@{}l}{\textit{Editing an existing repository}} \\
    SWE-bench & repository & issue & original tests & task & 33 lines (mean) & 2,294 & 1 \\
    SWE-Bench Pro & repository & issue, requirements, interface & original tests & task & 107 lines (mean) & 1,865 & 4 \\
    FeatureBench & repository or empty & statement, interfaces & original tests & task & 790 lines (mean, level 1) & 200 & 1 \\
    SWE-Milestone & earlier release & one document per milestone, by dependency & original tests & milestone & 570 lines (mean) & 98 milestones & 5 \\
    \midrule
    \multicolumn{8}{@{}l}{\textit{Building from scratch}} \\
    Commit0 & stubbed functions & documentation, visible tests & original tests & tests & \nr & 54 & 1 \\
    NL2Repo-Bench & empty & one document & original tests & tests & 1.5K to 4K (band of the median) & 104 & 1 \\
    BeyondSWE-Doc2Repo & empty & one specification & adapted original tests & tests & 3.5K (mean) & 50 & 1 \\
    E2E-SWE & empty & one specification & new tests & task & 7.4K (median) & 186 & 11 \\
    Zero2Repo & empty & requirements document, interface contract & rewritten original tests & task & \nr & 17 & 6 \\
    ProgramBench & executable & usage documentation & generated tests, partly from the original tests & task & 8.6K (median) & 200 & any \\
    MirrorCode & executable, visible tests & program behavior & end-to-end tests, partly from the original tests & task & 5.9K (median) & 25 targets & 6 \\
    PolyCodeEval L0 & empty & requirements document & interface-level functional tests & task, tests & 2.5K (mean) & 58 & 5 \\
    LEGO-REPO & interfaces, tests & public interface & original tests & tests & \nr & 522 & 1 \\
    E2EDev & empty & fine-grained requirements & generated tests, checked by people & requirement & \nr & 46 & web \\
    ICAE-Bench & empty & short request, user dialogue & converted and generated tests & tests & 35.5K (mean) & 480 & 12 \\
    RepoCraft & \nr & project description & original tests adapted by a language model & tests, coverage & 97.8K (mean) & 6 projects & 1 \\
    \midrule
    \textbf{NL2Repo-Pro} & empty, or starting files & overview document and one specification per feature, by dependency & original tests & feature & \pk{med-src-lines-k}K (median) & \pk{tasks-final} & 5 \\
    \bottomrule
  \end{tabular}
\end{table}

\section{Task Format and Release}
\label{app:format}

This appendix shows what a released task contains, first as excerpts of one task and then as the files of the release.

\subsection{A Sample Task}
\label{app:sample}

The excerpts come from the task built from \texttt{tobymao/sqlglot}, as the agent receives it. As in \Cref{sec:task-form}, task files label a feature \texttt{Capability $N$} and a dependency layer \texttt{Group $N$}. The overview document is cut to its section headings and one dependency layer, \texttt{Group 6}. The feature specification follows \texttt{Capability 30} of that layer, which has \pk{sample-fr} functional requirements and \pk{sample-ac} acceptance criteria, and keeps one of each. [\,\ldots\,] marks text left out, and the notes on the right name each part as in \Cref{sec:task-form}. The complete files are part of the released task set.

\begin{taskfile}{instruction.md}
\dochead{Build the \nolinkurl{sqlglot} library from this document}{}

\dochead{1. Overview}{what the library is}

This document specifies a Python package named \nolinkurl{sqlglot}. The workspace is empty. Build the package so that it provides every capability listed below and satisfies the Public Interface Contract.

\docskip
\dochead{2. Deliverable and environment}{deliverable and environment}

\docskip
\dochead{3. Ground rules}{rules for the whole library}

\begin{docitems}
\item Do not try to reach the network. The container has none. Use the installed dependencies recorded in \nolinkurl{/opt/nl2repo/constraints.txt}.
\end{docitems}

\docskip
\dochead{4. Capabilities}{features by dependency layer}

The capabilities are grouped by what they use, and the file of a capability names the files to read before it. A base class or helper a file uses may still be specified in a later file, so search \nolinkurl{/docs} for a name before you define it.

\docskip
\dochead{Group 6}{one dependency layer}

Complete statements can now enter through parsing or data modification builders, and separately supplied queries can be expanded into their callers. Building on expression composition, scope information, and inferred types, the rewrites reduce constant calculations, give equivalent expressions common forms, and discard intermediate columns that consumers do not need. This connects text input and query assembly with the forms used by later optimization and lineage analysis.

\begin{docitems}
\item \textbf{Capability 29. Parse SQL syntax into expression trees} in \nolinkurl{/docs/29-parse-sql-syntax-into-expression-trees.md}
\item \textbf{Capability 30. Simplify SQL expressions and fold constants} in \nolinkurl{/docs/30-simplify-sql-expressions-and-fold.md}
\item \textbf{Capability 31. Build data modification statements} in \nolinkurl{/docs/31-build-data-modification-statements.md}
\item \textbf{Capability 32. Expand named query sources into subqueries} in \nolinkurl{/docs/32-expand-named-query-sources-into.md}
\item \textbf{Capability 33. Canonicalize SQL expressions} in \nolinkurl{/docs/33-canonicalize-sql-expressions.md}
\item \textbf{Capability 34. Remove unused query projections} in \nolinkurl{/docs/34-remove-unused-query-projections.md}
\end{docitems}

\docskip
\dochead{5. Public Interface Contract}{summary of the public interface}

\nolinkurl{sqlglot.parse} accepts SQL and dialect options and returns statement trees in source order. \nolinkurl{Tokenizer.tokenize} gives access to tokens. In \nolinkurl{sqlglot.expressions.builders}, \nolinkurl{select} and \nolinkurl{from_} create queries that \nolinkurl{Select} and \nolinkurl{Query} methods extend with clauses and nested sources. Other builders create scalar expressions and definition or modification statements. \nolinkurl{sqlglot.expressions.convert} turns Python values into expressions, and \nolinkurl{DataType.build} accepts type text, a type category, or a type expression.

\docskip
\end{taskfile}

\begin{taskfile}{docs/30-simplify-sql-expressions-and-fold.md}
\dochead{Capability 30. Simplify SQL expressions and fold constants}{features to read first}

This capability uses what the capabilities below specify. Read them before this one.

\begin{docitems}
\item \textbf{Capability 1. Find and traverse SQL expression nodes} in \nolinkurl{/docs/01-find-and-traverse-sql-expression-nodes.md}
\item \textbf{Capability 4. Copy, compare, and hash SQL expression trees} in \nolinkurl{/docs/04-copy-compare-and-hash-sql-expression.md}
\item \textbf{Capability 12. Construct and inspect SQL data types} in \nolinkurl{/docs/12-construct-and-inspect-sql-data-types.md}
\end{docitems}

\docskip
\dochead{1. Background}{background}

SQL expression optimization needs to reduce computations and redundant conditions while preserving query meaning. This capability combines literal, boolean, temporal, and comparison simplification with canonical expression rendering for consistent comparison and ordering.

\dochead{2. Scope}{scope}

Implement expression-tree simplification and canonical expression rendering through the contracted interfaces, including constant folding, optional propagation and coalesce rewrites, dialect-sensitive behavior, and preservation of query structure.

\docskip
\dochead{3. Functional Requirements}{one of 20 requirements}

\docsub{FR-30.1}
Accept an expression tree and return its simplified expression. Update replacements within the supplied tree, including when the supplied expression is a child of a larger query, so callers retaining the enclosing query observe the change. Reapplying simplification must leave an already simplified result unchanged.

\docskip
\dochead{4. Public Interface Contract}{binding signatures, bodies left to the agent}

The following interface must be available under the stated modules, written as stubs: the signatures, decorators, and bases are normative, and every body is left to the implementation.

\docskip
\docsub{Module \nolinkurl{sqlglot.optimizer.simplify}}
\begin{docitems}
\item \textbf{Source path:} \nolinkurl{sqlglot/optimizer/simplify.py}
\end{docitems}

{\footnotesize
\begin{verbatim}
def simplify(
    expression: exp.Expr,
    constant_propagation: bool=False,
    coalesce_simplification: bool=False,
    dialect: DialectType=None
):
    ...
\end{verbatim}
}

\docskip
\dochead{5. Acceptance Criteria}{one of 15 criteria, with the requirements it covers}

\docsub{AC-30.1}
Given an enclosing projection containing the addition of 7 and 8, when its arithmetic child is simplified, then the enclosing projection contains the literal 15 and a second simplification makes no further change.

Requirements covered:

\begin{docitems}
\item FR-30.1
\item FR-30.3
\end{docitems}

\docskip
\end{taskfile}

\subsection{Release Format}
\label{app:release}

Each task is released as a Harbor task directory together with its two Docker images, one for the agent and one for the verifier.
\begin{description}
  \item[Instruction.] The overview document.
  \item[Environment.] The two images, with the pinned dependencies, and the files they were built from, including the feature specifications that the agent reads.
  \item[Tests.] Hidden from the agent. They hold the original tests, the graded list, the attribution of graded tests to features, and per feature its specification and core label.
  \item[Solution.] The original library code, which is the reference solution.
\end{description}
The verifier writes the test pass rate, FAR, the per-feature verdicts and the whole-task verdict for each run. These outputs are computed by program from the delivery and the released files. The score that Harbor aggregates over runs is FAR.

\FloatBarrier
\section{Construction}
\label{app:construction}

This appendix expands \Cref{sec:construction}. For each phase it describes what agents and programs produce and what reviewers check at the gate that closes it (\Cref{tab:gates}).

\begin{table}[!h]
  \centering
  \caption{Review gates. Every task is reviewed at every gate by engineers who work in its
  language.}
  \label{tab:gates}
  \scriptsize
  \begin{tabular}{@{}p{2.0cm}p{6.6cm}p{4.6cm}@{}}
    \toprule
    Gate & What reviewers check & What it secures \\
    \midrule
    Selection & License \newline Real use and a complete feature set \newline Executable tests that assert on main behavior \newline Pinned commit and excluded scope & A real library that can be evaluated \\
    \addlinespace[2pt]
    Environment & Reproducible rebuild \newline Tests exercise the delivered code \newline Every excluded test justified & Graded tests reflect the original library's behavior \\
    \addlinespace[2pt]
    Features & Each feature a coherent unit \newline Clear ownership of code and interfaces \newline Dependencies that follow the library's design & Features are meaningful units of functionality \\
    \addlinespace[2pt]
    Specifications & Requirements stated clearly and consistently \newline Graded behavior follows from the documents \newline No needless implementation constraints & Specifications can be answered, do not leak the original code, and admit any conforming implementation \\
    \addlinespace[2pt]
    Release & Pruning decisions \newline Main uses still graded after pruning & Fair and effective grading of the frozen version \\
    \bottomrule
  \end{tabular}
\end{table}

\paragraph{Selection.} Reviewers record the license of each repository (\Cref{app:per-task}). They admit a repository when it is a library in real use with a complete set of features, and when its main behavior is covered by executable tests that assert on behavior. Repositories whose main behavior needs network access, external services, special hardware or private data are excluded. Size is judged on effective source, module structure and functional complexity, with generated and vendored code left out. Each repository is pinned to a commit at which its code, documentation and tests agree.

\paragraph{An executable environment.} An agent installs the repository and determines which tests cannot run in the evaluation environment. It iterates until the original code passes every retained test and an empty workspace passes none. Running the tests also records which tests pass reliably, and the phase records which library code each test calls directly, from execution traces and call analysis (\Cref{app:attribution}). Reviewers check that the environment rebuilds reproducibly, that the tests exercise the delivered code, and that every excluded test has a reason.

\paragraph{Features and their dependencies.} A planning agent divides the library into features, each a unit of functionality that can be specified and accepted on its own. Several plans are drawn independently, and the one whose features cover the most code is kept. No piece of code belongs to more than one feature. The feature dependency graph is computed from the calls between features, and the features are grouped into dependency layers. A repository whose features do not depend on one another is dropped. Reviewers check that every feature is a coherent unit, that the ownership of code and interfaces is clear, and that the dependencies follow the library's design.

\paragraph{Specifications.} A specification agent writes one specification per feature with its background, scope, functional requirements and acceptance criteria. A program derives the interface section from the code the tests call directly. A feature specification is checked when the specification agent finishes it, and returned when a requirement disagrees with the interface or a behavior the tests rely on is left unstated. The overview document orders the feature specifications by dependency and adds a summary of the library, its interface and each layer. A skeleton keeps the reference solution's interface with every function body empty, and the original files that the verifier adds stay in place. Every test that passes on this skeleton is removed from grading. Reviewers check that the requirements are clear and consistent, that each graded behavior follows from the documents, and that the documents impose no needless implementation constraints.

\paragraph{Validation and pruning.} The task is packaged with its environment, its hidden tests and the original code as the reference solution, which must pass every graded test while an empty workspace scores zero. A judge agent then reads each graded test against the documents, without the original code, and removes from grading the tests whose behavior it finds the documents do not state. Reviewers check the pruning decisions and confirm that the main uses of the library are still graded.

\FloatBarrier
\section{The Task Set}
\label{app:per-task}

This appendix adds detail to \Cref{sec:task-set}: the original files that some tasks carry, a dependency from the same repository, the licenses, and every task with its results.

\paragraph{Starting files.} In \pk{start-tasks} tasks (\pk{start-rust} in Rust and \pk{start-java} in Java) the workspace starts with original files that lie outside the specification but that the build or the tests need. Most are test-support crates, procedural macro crates and entry points of binaries, and in several tasks they are empty or nearly empty. Over these tasks they amount to \pk{start-lines} non-blank lines, \pk{start-share}\% of their original source.

\paragraph{Added original files.} In \pk{placed-tasks} tasks the verifier adds original files when it grades a delivery, \pk{placed-entries} entries in all. Most are tests, test data, configuration and assets. The \pk{placed-nontest-src} source files outside test paths, in \pk{placed-nontest-src-tasks} tasks, are mostly scripts, build files, plugins and type declarations, and in JavaScript tasks also vendored third-party libraries. The agent sees their paths but not their contents, and they are in place when the skeleton check runs (\Cref{app:construction}).

\paragraph{Dependencies from the same repository.} The repository behind the \texttt{modelaudit} task holds several packages. The environment installs one of them, \texttt{modelaudit-picklescan}, from its published release as a declared dependency of the library. Its code overlaps with that of the graded package by \pk{modelaudit-overlap}\%, and \pk{modelaudit-tests} graded tests use it.

\paragraph{Licenses.} Each task keeps the license of its repository. Of the \pk{tasks} tasks, \pk{lic-permissive} come from repositories under permissive licenses, \pk{lic-strong-copyleft} under strong and \pk{lic-weak-copyleft} under weak copyleft licenses, \pk{lic-restrictive} under licenses that restrict commercial use, and \pk{lic-none} carry no license file.

\Cref{tab:tasks} lists every task.
{\scriptsize\setlength{\tabcolsep}{3pt}\begin{longtable}{@{}p{6.0cm}lcrrrrrrr@{}}
\caption{Every task: language, license class, graded features, graded tests, thousands of source lines, and FAR of each system. Lic.\ is P for permissive, W for weak copyleft, S for strong copyleft, R for restricted commercial use and N for no license file. Each task names its license in \texttt{task.toml}.}\label{tab:tasks}\\
\toprule
Repository & Lang. & Lic. & Feat. & Tests & Lines & \syslogo{claude} & \syslogo{gpt} & \syslogo{deepseek} & \syslogo{grok} \\
\midrule
\endfirsthead
\toprule
Repository & Lang. & Lic. & Feat. & Tests & Lines & \syslogo{claude} & \syslogo{gpt} & \syslogo{deepseek} & \syslogo{grok} \\
\midrule
\endhead
\bottomrule
\endlastfoot
\texttt{9600dev/mmr} & Python & R & 20 & 1,211 & 61 & 60.0 & 55.0 & 35.0 & 10.0 \\
\texttt{abundantbeing/hermes-browser-extension} & JavaScript & P & 124 & 912 & 116 & 82.3 & 58.1 & 55.6 & 57.3 \\
\texttt{AccumulateNetwork/accumulate} & Go & P & 37 & 528 & 212 & 13.5 & 16.2 & 8.1 & 8.1 \\
\texttt{AFCProject/AFC-Klipper-Add-On} & Python & S & 54 & 838 & 21 & 72.2 & 63.0 & 48.1 & 18.5 \\
\texttt{afonso360/ccc} & Rust & P & 27 & 310 & 98 & 29.6 & 37.0 & 22.2 & 25.9 \\
\texttt{agessaman/meshcore-bot} & Python & P & 110 & 1,646 & 83 & 36.4 & 77.3 & 30.0 & 19.1 \\
\texttt{aiming-lab/AutoResearchClaw} & Python & P & 159 & 1,271 & 77 & 98.1 & 14.5 & 86.8 & 84.3 \\
\texttt{air-embodied-brain/Zetta-Embodiment} & Python & N & 9 & 281 & 105 & 0.0 & 0.0 & 0.0 & 0.0 \\
\texttt{Alpha-Dojo/DojoAgents} & Python & P & 20 & 536 & 65 & 25.0 & 20.0 & 20.0 & 0.0 \\
\texttt{amathislab/DLC2action} & Python & S & 26 & 63 & 31 & 100.0 & 61.5 & 57.7 & 34.6 \\
\texttt{angr/angr-platforms} & Python & P & 21 & 22 & 32 & 100.0 & 38.1 & 38.1 & 0.0 \\
\texttt{apache/commons-jxpath} & Java & P & 33 & 310 & 28 & 81.8 & 45.5 & 27.3 & 15.2 \\
\texttt{ariekogan/ateam-mcp} & JavaScript & P & 15 & 90 & 15 & 73.3 & 73.3 & 26.7 & 53.3 \\
\texttt{ASomerN/node-red-contrib-octopus-intelligent} & JavaScript & P & 25 & 249 & 7 & 92.0 & 92.0 & 80.0 & 84.0 \\
\texttt{authzed/spicedb} & Go & P & 200 & 3,524 & 153 & 75.5 & 67.5 & 61.0 & 49.0 \\
\texttt{baileyh8/hermes-feishu-streaming-card} & Python & P & 9 & 475 & 68 & 44.4 & 55.6 & 11.1 & 0.0 \\
\texttt{BeamFour/Beam42} & Java & S & 71 & 224 & 58 & 50.7 & 50.7 & 31.0 & 15.5 \\
\texttt{beefsack/OmniTiler} & Rust & P & 29 & 1,076 & 127 & 55.2 & 58.6 & 55.2 & 24.1 \\
\texttt{benson/benson.github.io} & JavaScript & N & 51 & 306 & 44 & 92.2 & 94.1 & 66.7 & 3.9 \\
\texttt{blinklabs-io/gouroboros} & Go & P & 210 & 4,890 & 116 & 69.5 & 63.8 & 50.0 & 9.5 \\
\texttt{brycewang-stanford/StatsPAI} & Python & P & 74 & 3,790 & 370 & 60.8 & 47.3 & 41.9 & 27.0 \\
\texttt{caidaoli/ccLoad} & Go & P & 158 & 3,633 & 158 & 31.6 & 61.4 & 46.2 & 9.5 \\
\texttt{chakra-ui/panda} & Rust & P & 27 & 1,075 & 76 & 70.4 & 55.6 & 44.4 & 29.6 \\
\texttt{chemplexity/molecules} & JavaScript & P & 115 & 1,577 & 192 & 53.0 & 47.8 & 39.1 & 18.3 \\
\texttt{chengxianglibra/marivo} & Python & P & 79 & 1,419 & 148 & 46.8 & 41.8 & 27.8 & 11.4 \\
\texttt{cipherstash/sqltk} & Rust & P & 12 & 454 & 119 & 100.0 & 50.0 & 41.7 & 8.3 \\
\texttt{Cisco-Talos/EvidenceForge} & Python & P & 114 & 1,417 & 400 & 23.7 & 17.5 & 14.0 & 2.6 \\
\texttt{Clinical-Genomics/cg} & Python & N & 206 & 1,105 & 68 & 54.4 & 42.2 & 32.0 & 0.0 \\
\texttt{cmpute/dashu} & Rust & P & 68 & 995 & 67 & 82.4 & 73.5 & 25.0 & 8.8 \\
\texttt{cognitedata/cognite-sdk-python} & Python & P & 113 & 3,305 & 97 & 84.1 & 27.4 & 34.5 & 0.0 \\
\texttt{conradry/prtm} & Python & P & 6 & 19 & 89 & 66.7 & 33.3 & 33.3 & 33.3 \\
\texttt{cubed-dev/cubed} & Python & P & 75 & 955 & 12 & 80.0 & 36.0 & 60.0 & 5.3 \\
\texttt{CycloneDX/cyclonedx-core-java} & Java & P & 53 & 1,653 & 22 & 83.0 & 15.1 & 49.1 & 20.8 \\
\texttt{dan2097/opsin} & Java & P & 13 & 490 & 35 & 38.5 & 23.1 & 23.1 & 15.4 \\
\texttt{danielsobrado/drusniel-gods-end} & JavaScript & R & 104 & 370 & 56 & 91.3 & 80.8 & 76.0 & 58.7 \\
\texttt{denoptim-project/DENOPTIM} & Java & S & 26 & 233 & 96 & 42.3 & 19.2 & 15.4 & 11.5 \\
\texttt{devnied/EMV-NFC-Paycard-Enrollment} & Java & P & 43 & 294 & 25 & 90.7 & 53.5 & 37.2 & 30.2 \\
\texttt{digitalbazaar/forge} & JavaScript & P & 44 & 466 & 29 & 93.2 & 86.4 & 47.7 & 34.1 \\
\texttt{eakmanrq/sqlframe} & Python & P & 32 & 215 & 22 & 75.0 & 40.6 & 21.9 & 28.1 \\
\texttt{emmo-repo/EMMOntoPy} & Python & P & 26 & 44 & 12 & 84.6 & 80.8 & 73.1 & 23.1 \\
\texttt{FasterXML/woodstox} & Java & P & 76 & 843 & 55 & 94.7 & 69.7 & 50.0 & 32.9 \\
\texttt{feiskyer/koder} & Python & P & 21 & 1,330 & 80 & 0.0 & 9.5 & 9.5 & 0.0 \\
\texttt{fenecdb/fenec} & Rust & P & 53 & 308 & 82 & 47.2 & 39.6 & 15.1 & 9.4 \\
\texttt{flatppl/flatppl-rust} & Rust & P & 75 & 1,028 & 104 & 45.3 & 40.0 & 38.7 & 20.0 \\
\texttt{FreeOpcUa/opcua-asyncio} & Python & W & 48 & 596 & 326 & 47.9 & 50.0 & 0.0 & 0.0 \\
\texttt{FreeOpcUa/python-opcua} & Python & W & 20 & 275 & 217 & 85.0 & 20.0 & 15.0 & 0.0 \\
\texttt{freqtrade/freqtrade} & Python & S & 22 & 626 & 57 & 50.0 & 27.3 & 4.5 & 0.0 \\
\texttt{fuyuxiang/echo-agent} & Python & P & 294 & 2,749 & 86 & 73.1 & 32.0 & 50.0 & 20.7 \\
\texttt{generative-computing/mellea} & Python & P & 27 & 1,348 & 59 & 74.1 & 59.3 & 11.1 & 11.1 \\
\texttt{GhostInShells/MOSShell} & Python & P & 33 & 1,093 & 127 & 48.5 & 27.3 & 21.2 & 12.1 \\
\texttt{github/spec-kit} & Python & P & 7 & 834 & 54 & 71.4 & 42.9 & 14.3 & 14.3 \\
\texttt{globalpocket/brownie} & Rust & P & 11 & 169 & 167 & 27.3 & 45.5 & 27.3 & 27.3 \\
\texttt{Goldziher/ai-rulez} & Go & P & 194 & 5,916 & 213 & 10.3 & 44.8 & 38.7 & 7.2 \\
\texttt{grafana/gcx} & Go & P & 215 & 5,121 & 158 & 81.9 & 60.5 & 58.6 & 34.0 \\
\texttt{grelinfo/grelmicro} & Python & P & 26 & 1,737 & 77 & 23.1 & 30.8 & 7.7 & 0.0 \\
\texttt{GroophyLifefor/heap-analysis} & JavaScript & P & 15 & 118 & 2 & 93.3 & 6.7 & 93.3 & 93.3 \\
\texttt{gruntwork-io/terragrunt} & Go & P & 204 & 5,978 & 126 & 55.9 & 46.6 & 45.6 & 11.8 \\
\texttt{guaracloud/e-navigator} & Rust & P & 146 & 975 & 102 & 93.8 & 41.8 & 40.4 & 28.8 \\
\texttt{haowjy/meridian-cli} & Python & P & 29 & 674 & 99 & 27.6 & 20.7 & 10.3 & 13.8 \\
\texttt{huggingface/peft} & Python & P & 79 & 12,310 & 64 & 57.0 & 32.9 & 15.2 & 0.0 \\
\texttt{Intuition-Lab/personal-model} & Python & P & 118 & 959 & 58 & 68.6 & 56.8 & 42.4 & 19.5 \\
\texttt{iriumlabs/irium} & Rust & P & 37 & 667 & 187 & 81.1 & 54.1 & 70.3 & 35.1 \\
\texttt{italia/spid-cie-oidc-java} & Java & P & 25 & 164 & 5 & 100.0 & 64.0 & 84.0 & 56.0 \\
\texttt{jaymar921/HirayaCoder} & JavaScript & P & 57 & 571 & 32 & 77.2 & 73.7 & 73.7 & 54.4 \\
\texttt{jborean93/smbprotocol} & Python & P & 11 & 347 & 12 & 90.9 & 54.5 & 27.3 & 9.1 \\
\texttt{jhy/jsoup} & Java & P & 57 & 1,475 & 26 & 43.9 & 45.6 & 24.6 & 10.5 \\
\texttt{kapitalismho/PuriPuly-heart} & Python & S & 223 & 1,682 & 163 & 24.7 & 34.1 & 10.8 & 9.4 \\
\texttt{KashyapTan/Multplx} & Rust & P & 25 & 184 & 125 & 64.0 & 48.0 & 48.0 & 44.0 \\
\texttt{KittehOrg/KittehIRCClientLib} & Java & P & 55 & 230 & 32 & 100.0 & 72.7 & 54.5 & 60.0 \\
\texttt{Kozea/WeasyPrint} & Python & P & 20 & 550 & 29 & 75.0 & 60.0 & 45.0 & 40.0 \\
\texttt{kubernetes/apimachinery} & Go & P & 52 & 2,542 & 63 & 82.7 & 21.2 & 57.7 & 53.8 \\
\texttt{KumoCorp/shingetsu} & Rust & P & 69 & 1,297 & 86 & 47.8 & 43.5 & 29.0 & 17.4 \\
\texttt{kunifujiwara/VoxCity} & Python & P & 25 & 1,552 & 53 & 80.0 & 56.0 & 32.0 & 16.0 \\
\texttt{L1nze/LitBoard} & JavaScript & P & 21 & 127 & 63 & 81.0 & 90.5 & 81.0 & 61.9 \\
\texttt{lightningdevkit/rust-lightning} & Rust & P & 33 & 95 & 299 & 100.0 & 78.8 & 18.2 & 15.2 \\
\texttt{liiiiiiiiil/coding-agent-from-scratch} & Python & P & 43 & 247 & 32 & 51.2 & 53.5 & 20.9 & 16.3 \\
\texttt{Liu-Bot24/liusheng-subtitles} & JavaScript & P & 4 & 48 & 40 & 25.0 & 50.0 & 25.0 & 0.0 \\
\texttt{luceat-lux-vestra/arkst} & Rust & P & 36 & 802 & 107 & 58.3 & 61.1 & 44.4 & 16.7 \\
\texttt{Mathics3/mathics-core} & Python & S & 43 & 1,490 & 102 & 53.5 & 53.5 & 34.9 & 0.0 \\
\texttt{mattbasta/crass} & JavaScript & P & 5 & 225 & 6 & 80.0 & 80.0 & 80.0 & 80.0 \\
\texttt{menodata/time4j} & Java & W & 164 & 22,684 & 192 & 75.0 & 45.7 & 23.8 & 6.1 \\
\texttt{MetPX/sarracenia} & Python & S & 59 & 196 & 30 & 98.3 & 39.0 & 74.6 & 25.4 \\
\texttt{MichaelDurieuxInTenz/lire-compter-samuser} & JavaScript & N & 59 & 121 & 53 & 72.9 & 71.2 & 47.5 & 40.7 \\
\texttt{mikedh/trimesh} & Python & P & 109 & 541 & 48 & 77.1 & 0.9 & 24.8 & 0.9 \\
\texttt{mikera/vectorz} & Java & W & 38 & 342 & 53 & 13.2 & 18.4 & 13.2 & 5.3 \\
\texttt{mimno/Mallet} & Java & P & 27 & 157 & 90 & 48.1 & 55.6 & 14.8 & 37.0 \\
\texttt{mishka-zz/freecad-microwave} & Python & W & 98 & 3,137 & 92 & 2.0 & 23.5 & 14.3 & 3.1 \\
\texttt{modflowpy/flopy} & Python & P & 186 & 1,140 & 206 & 55.4 & 23.7 & 14.0 & 8.6 \\
\texttt{moleculer-java/moleculer-java} & Java & P & 28 & 101 & 25 & 71.4 & 35.7 & 46.4 & 25.0 \\
\texttt{MoSunDay/rdb} & Rust & P & 26 & 84 & 101 & 23.1 & 23.1 & 23.1 & 15.4 \\
\texttt{nadeemfareed/FAST-GC} & Python & S & 15 & 54 & 34 & 73.3 & 66.7 & 66.7 & 46.7 \\
\texttt{networknt/json-schema-validator} & Java & P & 22 & 858 & 25 & 50.0 & 54.5 & 36.4 & 18.2 \\
\texttt{NothingDNS/NothingDNS} & Go & P & 191 & 6,033 & 119 & 89.0 & 80.1 & 74.9 & 37.2 \\
\texttt{nschaetti/EchoTorch} & Python & S & 32 & 109 & 27 & 78.1 & 46.9 & 37.5 & 28.1 \\
\texttt{NygenAnalytics/scarf} & Python & P & 150 & 2,200 & 154 & 38.0 & 40.7 & 24.7 & 11.3 \\
\texttt{on4r4p/Chunklate} & Python & N & 100 & 979 & 79 & 62.0 & 25.0 & 52.0 & 47.0 \\
\texttt{Open-MSS/MSS} & Python & P & 85 & 390 & 72 & 0.0 & 10.6 & 32.9 & 20.0 \\
\texttt{open-policy-agent/opa} & Go & P & 134 & 3,677 & 145 & 53.7 & 47.0 & 39.6 & 29.9 \\
\texttt{openmagi/magi-agent} & Python & P & 9 & 486 & 368 & 33.3 & 33.3 & 33.3 & 22.2 \\
\texttt{openpnp/openpnp} & Java & S & 64 & 196 & 168 & 18.8 & 26.6 & 32.8 & 18.8 \\
\texttt{openstack/magnum} & Python & P & 19 & 578 & 18 & 57.9 & 0.0 & 0.0 & 0.0 \\
\texttt{opentypejs/opentype.js} & JavaScript & P & 14 & 110 & 18 & 100.0 & 78.6 & 57.1 & 50.0 \\
\texttt{pb33f/libopenapi} & Go & P & 14 & 2,291 & 69 & 28.6 & 42.9 & 14.3 & 0.0 \\
\texttt{pdm-project/pdm} & Python & P & 56 & 669 & 21 & 80.4 & 44.6 & 39.3 & 14.3 \\
\texttt{physwkim/rust-hdf5} & Rust & P & 47 & 297 & 93 & 2.1 & 31.9 & 8.5 & 2.1 \\
\texttt{pocketbase/pocketbase} & Go & P & 42 & 1,464 & 48 & 66.7 & 64.3 & 40.5 & 38.1 \\
\texttt{PoHsuanLai/porter} & Rust & P & 22 & 698 & 142 & 40.9 & 40.9 & 18.2 & 9.1 \\
\texttt{promptfoo/modelaudit} & Python & P & 21 & 1,008 & 237 & 38.1 & 42.9 & 19.0 & 9.5 \\
\texttt{pydcs/dcs} & Python & W & 48 & 72 & 177 & 87.5 & 0.0 & 29.2 & 0.0 \\
\texttt{r2dbc/r2dbc-mssql} & Java & P & 26 & 267 & 27 & 80.8 & 61.5 & 23.1 & 30.8 \\
\texttt{rackspace/runway} & Python & P & 26 & 923 & 59 & 96.2 & 61.5 & 34.6 & 30.8 \\
\texttt{Rimagination/scansci-harness} & Python & N & 155 & 644 & 106 & 16.8 & 56.1 & 52.9 & 37.4 \\
\texttt{rungalileo/galileo-python} & Python & P & 68 & 1,140 & 182 & 76.5 & 63.2 & 42.6 & 26.5 \\
\texttt{SafeExamBrowser/seb-server} & Java & W & 66 & 167 & 139 & 74.2 & 54.5 & 27.3 & 24.2 \\
\texttt{santiontanon/mdlz80optimizer} & Java & P & 21 & 158 & 43 & 9.5 & 14.3 & 14.3 & 14.3 \\
\texttt{scverse/scanpy} & Python & P & 43 & 1,454 & 39 & 76.7 & 55.8 & 37.2 & 16.3 \\
\texttt{seehuhn/go-pdf} & Go & S & 103 & 2,907 & 151 & 78.6 & 63.1 & 49.5 & 12.6 \\
\texttt{shisa-ai/hipEngine} & Python & S & 118 & 279 & 818 & 81.4 & 61.9 & 56.8 & 16.1 \\
\texttt{shopware/shopware-cli} & Go & P & 41 & 1,308 & 65 & 73.2 & 34.1 & 58.5 & 48.8 \\
\texttt{shreeshyla-hanasoge/Emberwick} & JavaScript & P & 21 & 136 & 24 & 61.9 & 76.2 & 47.6 & 23.8 \\
\texttt{skfolio/skfolio} & Python & P & 59 & 3,818 & 72 & 86.4 & 0.0 & 49.2 & 27.1 \\
\texttt{soklet/soklet} & Java & P & 48 & 593 & 64 & 54.2 & 37.5 & 29.2 & 18.8 \\
\texttt{Starainrt/astro} & Go & P & 137 & 1,912 & 99 & 58.4 & 28.5 & 24.8 & 11.7 \\
\texttt{sticktrk/rhythm-os} & Rust & P & 130 & 1,515 & 273 & 63.8 & 63.8 & 42.3 & 13.1 \\
\texttt{swaymun/wonder} & Rust & P & 50 & 125 & 107 & 78.0 & 56.0 & 70.0 & 20.0 \\
\texttt{tafia/quick-xml} & Rust & P & 11 & 1,113 & 33 & 72.7 & 54.5 & 18.2 & 0.0 \\
\texttt{tatuylonen/wiktextract} & Python & P & 28 & 1,178 & 809 & 14.3 & 10.7 & 10.7 & 7.1 \\
\texttt{telus-labs/stagecraft} & JavaScript & P & 44 & 468 & 51 & 93.2 & 86.4 & 84.1 & 61.4 \\
\texttt{tinygrad/tinygrad} & Python & P & 58 & 406 & 279 & 56.9 & 37.9 & 24.1 & 13.8 \\
\texttt{tobymao/sqlglot} & Python & P & 37 & 172 & 67 & 59.5 & 29.7 & 8.1 & 2.7 \\
\texttt{topcheer/ggcode} & Go & P & 98 & 5,724 & 435 & 49.0 & 41.8 & 41.8 & 28.6 \\
\texttt{translate/translate} & Python & S & 165 & 3,204 & 56 & 77.0 & 48.5 & 52.7 & 30.9 \\
\texttt{trpc-group/trpc-agent-go} & Go & P & 291 & 5,851 & 437 & 65.3 & 58.8 & 54.0 & 33.3 \\
\texttt{TSchonleber/brainctl} & Python & P & 91 & 858 & 87 & 83.5 & 82.4 & 70.3 & 28.6 \\
\texttt{twisted/pydoctor} & Python & P & 44 & 1,006 & 19 & 86.4 & 65.9 & 11.4 & 0.0 \\
\texttt{tyhts0829/grafix} & Python & P & 127 & 1,517 & 112 & 68.5 & 53.5 & 40.2 & 11.0 \\
\texttt{velocitykode/velocity} & Go & P & 239 & 3,864 & 135 & 72.0 & 61.9 & 56.1 & 39.3 \\
\texttt{Vheissu/emerald-bayou} & JavaScript & P & 82 & 202 & 28 & 84.1 & 87.8 & 74.4 & 51.2 \\
\texttt{viant/datly} & Go & P & 47 & 854 & 115 & 51.1 & 44.7 & 34.0 & 21.3 \\
\texttt{wakatime/wakatime-cli} & Go & P & 22 & 847 & 50 & 45.5 & 45.5 & 13.6 & 18.2 \\
\texttt{whad-team/whad-client} & Python & P & 30 & 801 & 80 & 76.7 & 66.7 & 50.0 & 36.7 \\
\texttt{wingbotai/wingbot} & JavaScript & P & 28 & 138 & 25 & 75.0 & 67.9 & 50.0 & 46.4 \\
\texttt{xhluca/session-migrate} & Python & P & 29 & 525 & 34 & 24.1 & 24.1 & 13.8 & 3.4 \\
\texttt{xiejhhhhhh/Draftpaper\_loop} & Python & R & 12 & 403 & 230 & 8.3 & 8.3 & 8.3 & 8.3 \\
\texttt{xiidoz/netnou} & JavaScript & P & 44 & 141 & 18 & 70.5 & 61.4 & 52.3 & 18.2 \\
\texttt{Yelp/paasta} & Python & P & 165 & 1,234 & 67 & 89.1 & 59.4 & 41.8 & 23.6 \\
\texttt{ymuromcev/ai-job-searcher} & JavaScript & P & 93 & 1,096 & 46 & 80.6 & 71.0 & 72.0 & 58.1 \\
\texttt{zenx0x/allinluna} & Python & P & 55 & 161 & 32 & 38.2 & 36.4 & 18.2 & 3.6 \\
\texttt{ZJU-PL/pyflow} & Python & P & 23 & 1,380 & 412 & 39.1 & 26.1 & 30.4 & 13.0 \\
\texttt{ZJunCher/mr-agent} & Python & S & 189 & 1,387 & 90 & 72.5 & 54.0 & 48.7 & 28.0 \\
\end{longtable}
}

\FloatBarrier
\section{Feature Attribution}
\label{app:attribution}

This appendix checks the rule that attributes graded tests to features (\Cref{sec:evaluation}) and how it bears on the results.

\paragraph{The rule.} A test counts for every feature whose code it calls directly (\Cref{sec:evaluation}). The attribution comes from execution traces and call analysis. A graded test is attributed to \pk{reach-mean} features on average, and \pk{reach-one}\% of graded tests to exactly one.

\paragraph{Audit.} When a task is built, the judge agent of the pruning step (\Cref{app:construction}) quotes the passage of the documents that states the behavior of each graded test. These quotes play no part in scoring, so they give an independent check of the attribution. We look up each quote in the documents and keep those that match exactly one document. Judge records exist for \pk{audit-tasks} tasks with \pk{audit-tests} graded tests. Among the tests whose quote matches exactly one document, that document is the specification of a feature the rule attributes the test to for \pk{audit-low}\% against the documents the judge read and \pk{audit-high}\% against the released documents.

\paragraph{Number of tests.} A feature is accepted only when all of its tests pass, so a feature with more tests must pass more cases. Within a task, each doubling of a feature's tests goes with a share of features accepted that is lower by \pk{opus-doubling} points for Claude Opus 5.5, \pk{sol-doubling} for GPT-6.1 Sol, \pk{deepseek-doubling} for DeepSeek V4.1 Flash and \pk{grok-doubling} for Grok 4.7. Core features carry more tests. Within a task their share accepted is lower by \pk{opus-core-gap}, \pk{sol-core-gap}, \pk{deepseek-core-gap} and \pk{grok-core-gap} points in the same order, and by \pk{opus-core-gap-ctrl}, \pk{sol-core-gap-ctrl}, \pk{deepseek-core-gap-ctrl} and \pk{grok-core-gap-ctrl} points once the number of tests is taken into account. Most of the gap for core features thus goes with their larger number of tests.

\paragraph{Dependency position.} To measure how acceptance falls along the dependency order (\Cref{sec:main-results}), we place each feature by its layer, from 0 for the first to 1 for the last, and fit the share accepted against this position within each task, so that only features of the same library are compared. A test that calls several features directly counts for all of them, which could affect this fall. Tests attributed to one feature are free of this effect. Using only those tests, the fitted fall from the first layer to the last is \pk{opus-depth-single}, \pk{sol-depth-single}, \pk{deepseek-depth-single} and \pk{grok-depth-single} points in the same order, so the rule alone does not produce it. Its size does depend on the rule.

\FloatBarrier
\section{Experimental Details}
\label{app:config}

This appendix gives the run settings and the measures behind \Cref{sec:experiments}, and compares FAR with the test pass rate for every pair of systems.

\begin{table}[!h]
  \centering
  \caption{Settings of every system. Context is the window given to the harness. The last
  column is the most subagents the harness runs at once. The agent phase has no network access
  in any run.}
  \label{tab:config}
  \scriptsize
  \setlength{\tabcolsep}{3pt}
  \begin{tabular}{@{}llllrr@{}}
\toprule
System & Harness & Model name & Reasoning effort & Context & Subagents at once \\
\midrule
\syslogo{claude}~Claude Opus 5.5 & Claude Code 2.1.311 & \texttt{claude-opus-5-5[1m]} & xhigh & 1,000,000 & 20 \\
\syslogo{gpt}~GPT-6.1 Sol & Codex CLI 0.164.0 & \texttt{gpt-6.1-sol} & xhigh & 1,050,000 & 3 \\
\syslogo{deepseek}~DeepSeek V4.1 Flash & Claude Code 2.1.311 & \texttt{deepseek-v4.1-flash} & xhigh & 1,000,000 & 20 \\
\syslogo{grok}~Grok 4.7 & Claude Code 2.1.311 & \texttt{grok-4.7} & xhigh & 490,000 & 20 \\
\bottomrule
\end{tabular}

\end{table}

\paragraph{Settings.} \Cref{tab:config} lists the harness and model settings of every system.

\paragraph{Resource measures.} A run consists of a main agent and the subagents it spawns, including subagents spawned by subagents. A request is one call to the model, and a tool call is one call to a tool. Shell commands are classified by purpose, such as running tests, building, editing, searching and reading. A command can fall into several classes. Input tokens are summed over all requests of a run, so context that is sent again counts each time. The context peak is the largest input of a single request by one agent. Context compaction is rare. The context of the main agent is compacted in at most \pk{comp-max} of \pk{tasks} runs for any system, so nearly every run keeps its full history.

\paragraph{Order measures.} These measures use the writes that can be read from a trajectory. Between \pk{ord-unread-min}\% and \pk{ord-unread-max}\% of writes cannot be read, for instance code produced by a generator script, and these are not attributed to features. For dependency edges, the first implementation of a feature is the first non-stub line that carries two or more names from its interface that distinguish it from other features, or one such name in the file the original library uses for it. The shuffled baseline permutes the first-implementation times among the features of the same run and takes the same share of edges. The layer measures accept any such name, which can place a first implementation earlier. Between two layers, any test command counts, whatever its outcome. Reading times come from the feature specification files that a trajectory shows being opened. Across systems, the median Spearman correlation between the layer of a feature and the time its specification is first read ranges from \pk{ord-read-min} to \pk{ord-read-max}. Test commands run through most of a run, with \pk{ord-first-min}\% to \pk{ord-first-max}\% of them in the first tenth and \pk{ord-last-min}\% to \pk{ord-last-max}\% in the last, against 10\% for an even spread. Feature specifications are numbered by layer, so the order in which features are built may also come from following their numbers. Within each system, the share of edges built in order is not correlated with FAR, with a Spearman correlation between \pk{ord-far-min} and \pk{ord-far-max}.

\paragraph{Behaviors.} Software engineers label each run in the main results for 23 behaviors as present, absent or unclear, and every present label carries a quote from the trajectory. The behaviors cover how the run starts, how it uses subagents, how it relates to the original library, how it verifies its work and how it delivers. Behaviors that presuppose subagents are counted only in runs that spawn them. Percentages are taken over the runs in which a label could be decided.

\paragraph{Variation between runs.} A few runs account for much of the variation of GPT-6.1 Sol. Runs more than 20 points below the median of their task make up \pk{collapsed-runs-pct}\% of runs and \pk{collapsed-share}\% of the variation within tasks, and most of them did not build or load completely. These runs also drive the relation between resource use and score (\Cref{sec:work}). Across the five runs on each task, FAR rises by \pk{sol5-out-slope} points per doubling of output tokens (95\% interval \pk{sol5-out-lo} to \pk{sol5-out-hi}), but most of the rise comes from these \pk{sol5-collapsed} of the \pk{sol5-runs} runs. Without them, FAR rises by \pk{sol5-out-slope-nc} points per doubling (95\% interval \pk{sol5-out-lo-nc} to \pk{sol5-out-hi-nc}), an interval that includes no change. Cumulative input and tool calls show no rise even with these runs included.

\paragraph{FAR and test pass rate.} FAR gives the larger effect in every pair of systems (\Cref{tab:pairs}), by a wide margin in the three pairs with Claude Opus 5.5 and by a small one in the other three.

\begin{table}[t]
  \centering
  \caption{Differences between systems on FAR and on the test pass rate, paired by task. A gap is the difference between the scores of System A and System B in \Cref{tab:main}, in points, which equals their mean per-task difference up to rounding. An effect divides that mean by the standard deviation of the per-task differences, so it shows how consistently System A scores above System B.}
  \label{tab:pairs}
  \small
  \begin{tabular}{@{}llrrrr@{}}
\toprule
System A & System B & FAR gap & FAR effect & Pass-rate gap & Pass-rate effect \\
\midrule
\syslogo{claude}~Claude Opus 5.5 & \syslogo{gpt}~GPT-6.1 Sol & 14.6 & 0.61 & 5.4 & 0.22 \\
\syslogo{claude}~Claude Opus 5.5 & \syslogo{deepseek}~DeepSeek V4.1 Flash & 24.0 & 1.13 & 13.7 & 0.64 \\
\syslogo{claude}~Claude Opus 5.5 & \syslogo{grok}~Grok 4.7 & 39.2 & 1.63 & 35.8 & 1.23 \\
\syslogo{gpt}~GPT-6.1 Sol & \syslogo{deepseek}~DeepSeek V4.1 Flash & 9.4 & 0.45 & 8.3 & 0.36 \\
\syslogo{gpt}~GPT-6.1 Sol & \syslogo{grok}~Grok 4.7 & 24.6 & 1.11 & 30.4 & 1.10 \\
\syslogo{deepseek}~DeepSeek V4.1 Flash & \syslogo{grok}~Grok 4.7 & 15.2 & 1.09 & 22.1 & 0.96 \\
\bottomrule
\end{tabular}

\end{table}

\FloatBarrier
\section{Line Overlap and Recall Measures}
\label{app:recall}

This appendix defines the measures in \Cref{sec:recall}: line overlap, hidden test names and statements of recall.

\paragraph{Line overlap.} We compare the library code of each delivery with that of the original, leaving out test directories. Both sides are first formatted with the default settings of the same standard tool: black for Python, gofmt for Go, rustfmt for Rust, google-java-format for Java and prettier for JavaScript. Files the tool cannot process, \pk{orig-unformatted}\% of the original files, stay as they are. A code line has at least 25 characters after whitespace normalization and is not a comment, an import or package declaration, a lone bracket or a similar trivial line. Lines identical to a line of the task documents or the starting files are left out on both sides. Line overlap is the share of the code lines of a delivery, counted with repetition, whose text matches some line of the original, in any file and at any position. The baseline in \Cref{sec:recall} is this share on repositories created in 2026.

\begin{table}[t]
  \centering
  \caption{Line overlap before and after formatting, and the share of token 5-grams of a delivery
  that occur in the original, by the year the repository was created. Medians over runs, in percent.}
  \label{tab:overlap}
  \scriptsize
  \setlength{\tabcolsep}{3pt}
  \begin{tabular}{@{}lr*{12}{>{\raggedleft\arraybackslash}p{2.3em}}@{}}
\toprule
& & \multicolumn{4}{c}{Line overlap, unformatted} & \multicolumn{4}{c}{Line overlap, formatted} & \multicolumn{4}{c}{Token 5-grams} \\
\cmidrule(lr){3-6}\cmidrule(lr){7-10}\cmidrule(l){11-14}
Created & Tasks & \syslogo{claude} & \syslogo{gpt} & \syslogo{deepseek} & \syslogo{grok} & \syslogo{claude} & \syslogo{gpt} & \syslogo{deepseek} & \syslogo{grok} & \syslogo{claude} & \syslogo{gpt} & \syslogo{deepseek} & \syslogo{grok} \\
\midrule
2020 or earlier & 59 & 44.0 & 3.8 & 16.2 & 17.4 & 50.4 & 14.6 & 16.9 & 19.2 & 58.9 & 26.0 & 29.0 & 28.6 \\
2021 to 2023 & 22 & 28.7 & 5.5 & 15.9 & 18.2 & 30.0 & 11.8 & 16.8 & 19.0 & 44.5 & 31.8 & 37.1 & 35.4 \\
2024 to 2025 & 25 & 16.9 & 3.7 & 14.3 & 17.2 & 18.1 & 14.8 & 15.1 & 18.8 & 32.4 & 26.1 & 27.0 & 29.2 \\
2026 & 44 & 12.7 & 2.2 & 14.0 & 16.8 & 15.1 & 14.5 & 15.2 & 18.1 & 28.5 & 26.7 & 26.8 & 31.4 \\
\bottomrule
\end{tabular}

\end{table}

\paragraph{Layout and the specification.} Formatting matters most for GPT-6.1 Sol, whose unformatted lines average \pk{sol-cpl} characters, as a median over runs, against \pk{orig-cpl} in the original libraries. On repositories created in 2026, its overlap is \pk{sol-base-raw}\% before formatting and \pk{sol-base}\% after, while the other systems change by a few points (\Cref{tab:overlap}). Token 5-gram overlap does not depend on layout. It splits the code into tokens without comments, commas, semicolons and quotes, and takes the share of the token 5-grams of a delivery that occur in the original. Between the repositories created in 2026 and those from 2020 or earlier, it rises from \pk{opus-grams-new}\% to \pk{opus-grams-old}\% for Claude Opus 5.5 and changes by a few points at most for the other systems. Some matching lines come from the specification itself. On repositories created in 2026, \pk{spec-lines-min}\% to \pk{spec-lines-max}\% of the matching lines before formatting have their full text inside a line of the task documents, for instance an interface declaration given in running text. Without them, the unformatted overlap there is \pk{opus-base-nospec}\%, \pk{sol-base-nospec}\%, \pk{deepseek-base-nospec}\% and \pk{grok-base-nospec}\% for Claude Opus 5.5, GPT-6.1 Sol, DeepSeek V4.1 Flash and Grok 4.7.

\paragraph{Feature-level overlap.} The comparison of features in \Cref{sec:recall} uses a separate measure. The original code of a feature is the set of original source files that define the names in its interface section, leaving out names defined in more than three files. A file can serve several features, so this mapping is coarse. The overlap of a feature is the share of the code lines of these files that a run writes verbatim, with whitespace normalized and without formatting. Unlike line overlap, it counts lines of the original rather than of the delivery. We keep the \pk{fov-features} features with at least ten such lines. Within each task, we sort the features by the overlap of Claude Opus 5.5 and compare the upper half with the lower half, on the \pk{fov-tasks} tasks with at least three features in each half. The first run of GPT-6.1 Sol on the same features serves as a control, since its difference between the halves reflects how hard these features are. On average per task, Claude Opus 5.5 has \pk{fov-opus-high}\% of the upper half and \pk{fov-opus-low}\% of the lower half accepted, and GPT-6.1 Sol \pk{fov-sol-high}\% and \pk{fov-sol-low}\%. The difference for Claude Opus 5.5 minus that for GPT-6.1 Sol is \pk{fov-did} points (95\% interval \pk{fov-did-lo} to \pk{fov-did-hi}). Across tasks, \pk{fov-high-old}\% of the features where Claude Opus 5.5 writes at least half of the original code come from repositories from 2020 or earlier, where its lead is larger, so the comparison stays within tasks.

\paragraph{Hidden test names.} We count the names of hidden tests that a run writes in its tool inputs. Names must have at least 12 characters, or 20 for JavaScript test titles, and must not appear in the task documents. By the behavior labels of \Cref{app:config}, Claude Opus 5.5 also rewrites original tests from memory to check its own work in \pk{b-opus-recalled-upstream-tests}\% of runs.

\paragraph{Statements of recall.} We count first-person statements about rebuilding the original from memory in the reasoning of each run, and classify each hit by a window of about 200 characters around it. Reconstruct states an intent to write the code from memory. Recall original states a memory of the project without such an intent. Recall public refers to public standards or algorithms, and Negated says the model does not remember. The measure in \Cref{tab:age}, stated intent to rebuild, is the share of runs with a Reconstruct hit. \Cref{tab:said} gives the counts. These are word-based counts, so we use them for trends within a system. The reasoning of GPT-6.1 Sol is encrypted, and that of Claude Opus 5.5 is a summary. Only for Claude Opus 5.5 does the number of statements of intent per character of reasoning go with line overlap within its runs, with a Spearman correlation of \pk{opus-said-overlap}, against \pk{deepseek-said-overlap} for DeepSeek V4.1 Flash and \pk{grok-said-overlap} for Grok 4.7.

\begin{table}[t]
  \centering
  \caption{Statements in the reasoning of each system. Any statement counts runs with a Reconstruct or Recall original hit, out of the runs whose trajectory keeps reasoning text. For Claude Opus 5.5 these are \pk{opus-said-runs} of its \pk{tasks} runs. The other cells give hits and the runs they occur in.}
  \label{tab:said}
  \small
  \begin{tabular}{@{}lrrrrr@{}}
\toprule
System & Any statement & Reconstruct & Recall original & Recall public & Negated \\
\midrule
\syslogo{claude}~Claude Opus 5.5 & 115/144 & 5,284 / 99 & 18,588 / 115 & 1,140 / 83 & 2,550 / 105 \\
\syslogo{deepseek}~DeepSeek V4.1 Flash & 135/150 & 12,125 / 123 & 36,004 / 132 & 929 / 84 & 2,650 / 110 \\
\syslogo{grok}~Grok 4.7 & 97/150 & 663 / 81 & 1,543 / 91 & 85 / 40 & 17 / 14 \\
\bottomrule
\end{tabular}

\end{table}

\end{document}